\documentclass[twocolumn,10pt]{article}

\title{Bayesian Optimisation for Mixed-Variable Inputs using Value Proposals}

\usepackage{microtype}
\usepackage{graphicx}
\usepackage{subfig}
\usepackage{caption}
\usepackage{booktabs} 

\usepackage{xspace}
\usepackage{acronym}
\usepackage{amsfonts}
\usepackage{amsthm}
\usepackage{standalone}
\usepackage{bm}
\usepackage{algorithm}
\usepackage{algpseudocode}
\usepackage{cite}
\usepackage{makecell}
\usepackage{courier}
\usepackage{verbatim}
\usepackage{multirow}
\usepackage{amsmath,amssymb}
\usepackage{xcolor}
\usepackage{enumitem}
\usepackage{setspace}
\usepackage{ulem}
\usepackage{authblk}

\usepackage{hyperref}

\makeatletter
\DeclareRobustCommand\onedot{\futurelet\@let@token\@onedot}
\def\@onedot{\ifx\@let@token.\else.\null\fi\xspace}

\def\eg{\textit{e.g}\onedot} 
\def\ie{\textit{i.e}\onedot} 
 
\def\etc{\textit{etc}\onedot}

\makeatother

\newcommand{\matern}{Mat\'{e}rn}
\DeclareMathOperator*{\argmax}{arg max}

\algnewcommand\algorithmicforeach{\textbf{for each}}
\algdef{S}[FOR]{ForEach}[1]{\algorithmicforeach\ #1\ \algorithmicdo}

\newtheorem{theorem}{Theorem}
\newtheorem{asu}{Assumption}
\newtheorem{definition}{Definition}
\newtheorem{lemma}{Lemma}

\author[1,2]{Yan Zuo}
\author[1]{Amir Dezfouli}
\author[1]{Iadine Chad\`{e}s}
\author[1]{David Alexander}
\author[2]{Benjamin Ward Muir}

\affil[1]{Data61, CSIRO}
\affil[2]{Manufacturing, CSIRO}

\begin{document}

\maketitle

\begin{abstract}
Many real-world optimisation problems are defined over both categorical and continuous variables, yet efficient optimisation methods such as Bayesian Optimisation (BO) are not designed to handle such mixed-variable search spaces. Recent approaches to this problem cast the selection of the categorical variables as a bandit problem, operating independently alongside a BO component which optimises the continuous variables. In this paper, we adopt a holistic view and aim to consolidate optimisation of the categorical and continuous sub-spaces under a single acquisition metric. We derive candidates from the Expected Improvement criterion, which we call value proposals, and use these proposals to make selections on both the categorical and continuous components of the input. We show that this unified approach significantly outperforms existing mixed-variable optimisation approaches across several mixed-variable black-box optimisation tasks.
\end{abstract}

\section{Introduction}
\label{sec:intro}
Bayesian optimisation (BO) has established itself as an efficient method for optimising black-box functions that are costly to evaluate~\cite{shahriari2015taking}. Typical BO methods model the black-box function of interest using a surrogate statistical model (usually a Gaussian Process (GP)~\cite{dudley2010sample}), seeking out the next point to evaluate by optimising a more tractable (typically differentiable) function called an acquisition function. The role of the acquisition function is to balance two conflicting requirements: exploitation of the current knowledge about the objective function and exploration to gain more knowledge about the objective function. BO has been applied effectively to tasks that range from experimental design~\cite{griffiths2017constrained, shields2021bayesian} to hyperparameter search~\cite{snoek2012practical,gardner2014bayesian} in machine learning models. Notably, it is often observed that BO is particularly well-suited for applications where the number of allowable evaluations on the objective function is limited~\cite{bull2011convergence}.

However, many real world optimisation problems involve a mixture of continuous and categorical variables in the input space. For example, in automated machine learning applications~\cite{hutter2019automated} where the aim is to automatically select a machine learning model along with its corresponding optimal hyperparameters, each model can be seen as a categorical choice while the hyperparameters of the model can be viewed as category-specific continuous variables. Another example is in the chemical reaction space, where often the function we are interested in optimising is represented by both categorical (compositional) variables and continuous (process) variables~\cite{zhou2017optimizing}. These scenarios present an additional challenge for current BO models (particularly those using GPs as their underlying surrogate models) which are ill-equipped to deal effectively with such problems containing multi-layered and complicated search spaces. 

There are numerous challenges associated with optimising mixed-variable functions within a BO framework. In particular, the assumption that the acquisition function is differentiable over the input space (allowing for efficient optimisation) becomes invalid. Recent works have dealt with the categorical part of the input in various ways ranging from one-hot encoding of the categorical components~\cite{snoek2012practical, golovin2017google} to using hierarchical models~\cite{hutter2011sequential, bergstra2011algorithms} suited towards both categorical and continuous inputs. Several modern approaches for optimising mixed-variable functions utilise a mixed approach involving Multi-Armed Bandits (MABs)~\cite{auer2002finite} and BO for handling the categorical and continuous parts respectively. However, these methods either treat the mixed-variable optimisation problem as a collection of smaller continuous problems~\cite{gopakumar2018algorithmic,nguyen2020bayesian} which is sample-inefficient, or separately optimise for each categorical variable~\cite{ru2020bayesian} which is a non-global approach. 


In this paper, we present a new Bayesian Optimisation approach for optimising black-box functions with multiple continuous and categorical inputs. For carrying out a decision on the categorical part of the input, our method uses a set of acquisition values derived from our surrogate model, which we call \textit{value proposals}. This enables a global, unified approach for optimising on the discrete and continuous sub-spaces of the input, where the decision-making process for both categorical and continuous variables is based on a common metric obtained from the underlying surrogate. Bayesian Optimisation using Value Proposals (VPBO) offers the following main contributions:
\begin{itemize}
\item We propose a novel BO approach for optimising mixed-variable black-box functions using value proposals, unifying the joint optimisation of categorical and continuous inputs under a single framework.
\item We derive the regret bound for VPBO, showing that the regret for our method grows only sub-linearly in the number of iterations. 
\item We show our method significantly improves over existing baselines when applied to a variety of mixed input synthetic and real-world problems.
\end{itemize}

\section{Background}
\label{sec:background}
Generally, we can organise the literature related to our work into three categories:

\paragraph{One-hot Encoding}
Prior to the introduction of hierarchical and categorical-specific models for mixed input tasks, one-hot encoding methods~\cite{snoek2012practical, gonzalez2016batch, golovin2017google}, which transform categorical variables into a one-hot encoded representation, were used by Bayesian Optimisation frameworks to deal with inputs of mixed nature. In this scenario, the categorical variable with $N$ choices is transformed into a vector of length $N$ with a single non-zero element. Since categories are mutually exclusive, this type of approach treats each extra variable as continuous in $[0,1]$ using a standard Bayesian Optimisation algorithm for optimisation. However, this type of approach places an equal measure of covariance between all category pairs (despite some or all pairs having different or no correlations), resulting in an acquisition function that is difficult to optimise with large areas of flatness~\cite{rana2017high}. To address this issue,~\cite{garrido2020dealing} restricted the objective function to change only at designated points of $0$ and $1$, using a kernel function which computed covariances after rounding off the input. However, with this approach, the resulting acquisition function becomes step-wise, making it difficult to optimise.

\paragraph{Hierarchical}
Another approach to dealing with mixed inputs is to change the underlying surrogate model for the black-box function from the commonly used Gaussian Process to a model that can more naturally consider both continuous and categorical variables. Sequential Model-based Algorithm Configuration (SMAC)~\cite{hutter2011sequential} uses Random Forests (RFs)~\cite{breiman2001random} as a surrogate model to handle both categorical and continuous components in the input. However, the random nature of RFs (through a reliance on bootstrapping samples and randomly choosing subsets of variables to be tested at each node) weakens the reliability of the derived acquisition function. Adding to this, RFs have a tendency to overfit to training data, requiring careful selection of the number of trees in the model to avoid overfitting. Other tree-based approaches include the Tree-Parzen Estimator~\cite{bergstra2011algorithms}, which uses tree-structured Parzen density estimators.

\paragraph{Category-specific}
The category-specific approach is to handle each component of the input separately.~\cite{gopakumar2018algorithmic} developed EXP3BO, an approach to deal with mixed categorical and continuous input spaces by utilising a Multi-Armed Bandit to make categorical choices and training a separate surrogate model specific for each choice of category. As a result, the observed data is divided into smaller subsets (one for each category), resulting in a sample-inefficient optimisation procedure that cannot handle problems with a large number of categorical choices.~\cite{nguyen2020bayesian} introduced a batched setting to the optimisation framework of~\cite{gopakumar2018algorithmic} and replaced the respective EXP3 and Upper Confidence Bound (UCB) algorithms of the framework with Thompson Sampling~\cite{chapelle2011empirical}. A key limitation of these frameworks is that they only allow for optimisation of a single categorical variable; the work of~\cite{ru2020bayesian} extended this type of approach by allocating a MAB per categorical variable, enabling the optimisation of functions with multiple categorical variables. However, each MAB is individually updated using the EXP3 algorithm and this may lead to estimates which are disjoint from the BO-backend. Our method builds on these hybrid approaches but instead adopts a global view for selecting categorical combinations in the discrete sub-space of the input.

\section{Preliminaries}
\label{sec:prelims}

\subsection{Problem Setup}
We consider the problem of optimising a black-box function $f(\bm{z})$ where the input $\bm{z}$ is comprised of categorical and continuous parts \ie $\bm{z} = [\bm{h}, \bm{x}]$. Here, $\bm{h} = [h_1, ..., h_k]$ is a vector of categorical variables from the discrete topological space $\mathcal{C}$, with each categorical variable $h_i \in \{1,2,..., N_j\}$ taking one of $N_j$ different values. 
The continuous component of the input, $\bm{x}$, is drawn from a $d_x$-dimensional hypercube $\mathcal{X}$. 
Formally, the optimisation of the black-box function $f$ is expressed as:
\begin{equation}
    \bm{z}^* = [\bm{h}^*, \bm{x}^*] = \argmax_z f(\bm{z})
    \label{eq:main_opt}
\end{equation}
which is performed sequentially by making a series of evaluations on $z_1, ...z_T$. The goal is to find the best configuration $\bm{z}^*$ that maximises $y$, which is the value returned from our objective function $f$. For convenience, we use $c$ to denote a possible combination of categorical choices out of $C = \prod_{j = 1}^{k}{N_j}$ available combinations, such that $c \in \{1,...,C\}$.

\subsection{Bayesian Optimisation}
\label{ssec:bayesian_optimisation}
Given a black-box objective function $f : \mathcal{X} \rightarrow \mathbb{R}$, the goal of BO is to find the optimal value $\bm{x}^*$ under a setting with limited evaluations on $f$. This optimal value maximises the objective $f$, and is defined as $\bm{x}^* = \argmax_{\bm{x} \in \mathcal{X}} f(\bm{x})$. The process of BO involves using a surrogate to model the objective $f$; typically, $f$ is assumed to be a smooth function and commonly a Gaussian Process (GP) is used for the surrogate. The GP models an underlying probability distribution over functions $f$ and is represented by mean and covariance functions (or kernel) $\mu(\bm{x})$ and $\kappa(\bm{x}, \bm{x}')$ respectively, where $f(\bm{x}) \sim \mathtt{GP}(\mu(\bm{x}), \kappa(\bm{x}, \bm{x}'))$. 


The surrogate encodes our prior beliefs about the objective $f$ and we can build a posterior through further observations when we evaluate $f(\bm{x})$. Using this posterior, at a given optimisation iteration $t$, an acquisition function $\alpha_t(\bm{x})$ can be built, which can then be optimised to identify the next point to be sampled such that $\bm{x}_t = \argmax_{x \in \mathcal{X}} \alpha_t(\bm{x})$. Since $\alpha_t(\bm{x})$ is derived from our surrogate model, it is comparatively cheaper to compute and can be optimised using standard optimisation techniques.

\subsection{Expected Improvement}
\label{ssec:expected_improvement}
For our choice of acquisition function, we use Expected Improvement (EI), which is an expectation over the improvement function. The improvement function at iteration $t$ is given by:
\begin{equation}
    \mathcal{I}_t(\bm{x}_t) = \max{} \{0, f(\bm{x}_t) - \mathcal{E}_t\},
\label{eq:improvement_function}
\end{equation}
where $\mathcal{E}_t$ defines an incumbent at iteration $t$, such that $\mathcal{E}_t = y_{t-1}^{max} = \max{} \{y_1, ...y_{t - 1} \}$. We denote $\gamma = \gamma_{t-1}(\bm{x}) = \frac{\mu_{t-1}(\bm{x}) - \mathcal{E}_t}{\sigma_{t-1}(\bm{x})}$, where $\mu_{t-1}(\bm{x})$ and $\sigma^2_{t-1}(\bm{x})$ are respectively the predictive mean and variance of the posterior. Then we can obtain the closed-form acquisition function at iteration $t$ by taking the expectation over Eq.~\ref{eq:improvement_function} (refer to Appendix~\ref{app:EI_closed_form} for derivation):
\begin{align}
\notag \alpha_t^{EI}(\bm{x}) &= \mathbb{E}[\mathcal{I}_t(\bm{x})] \\
&= \sigma_{t-1}(\bm{x})\phi(\gamma) + (\mu_{t-1}(\bm{x}) - \mathcal{E})\Phi(\gamma),
\label{eq:expected_improvement}    
\end{align}
where $\phi$ and $\Phi$ are the PDF and CDF of the standard normal distribution respectively. In the case $\sigma_{t-1}(\bm{x}) = 0$, we define $\alpha_t^{EI}(\bm{x}) = 0$.



\subsection{Mixed-Kernels}
Under the typical setting of BO, where the surrogate is defined by a Gaussian Process, the GP kernel is defined over a continuous space. This makes using GP-based BO approaches on mixed-variable problems especially difficult. We adopt the mixed-kernel approach of~\cite{ru2020bayesian} which defines two separate kernels: a categorical kernel $\kappa_h(\bm{h}, \bm{h}')$ defined over the discrete sub-space and a continuous kernel $\kappa_x(\bm{x}, \bm{x}')$ over the continuous sub-space. The mixed-kernel $\kappa_z(\bm{z}, \bm{z}')$ is then defined as a mixture of sum and product kernels over $\kappa_h$ and $\kappa_x$:
\begin{align}
    \notag \kappa_z(\bm{z}, \bm{z}') &= (1-\lambda)(\kappa_h(\bm{h}, \bm{h}')+\kappa_x(\bm{x}, \bm{x}'))
    \\ &+ \lambda \kappa_h(\bm{h}, \bm{h}')\kappa_x(\bm{x}, \bm{x}'),
\end{align}
with $\lambda \in [0,1]$ being a trade-off parameter for the sum and product kernels which can be optimised jointly alongside the GP hyperparameters. The mixed-kernel is able to learn covariances between observations made in the joint input space $\mathcal{Z} = \mathcal{C} \times \mathcal{X}$. Samples from different categories will have covariances which are dominated by $\kappa_x$ and when there is partial overlap in categorical choices between samples, the covariance is determined by contributions from both $\kappa_h$ and $\kappa_x$ (for additional details on the mixed-kernel, refer to the Appendix~\ref{app:mixed_kernel}).

\section{Method}
\label{sec:method}


\subsection{Bayesian Optimisation using Value Proposals}

Our proposed method, Bayesian Optimisation using Value Proposals (VPBO), extends hybrid MAB-BO approaches such as~\cite{gopakumar2018algorithmic, nguyen2020bayesian, ru2020bayesian} to adopt a global view for optimising mixed-variable input spaces. This allows our method to retain the advantages of bandit approaches for discrete-type problems such as selecting categorical inputs. 

The VPBO procedure first queries the BO back-end by maximising the acquisition function to make a selection on the continuous part of the input $\bm{x}_t$. This step is done for each of the $C$ possible combinations of categories. For each combination of categories $c$, VPBO generates a \textit{value proposal}, which is collected in the set of all value proposals $\mathcal{V}_t$ for the optimisation iteration $t$, along with their corresponding $\bm{x}^*_{t,c}$. Deciding the categorical choices $\bm{h}^*_t$ at iteration $t$ is then formulated as a MAB problem~\cite{bubeck2008online, vermorel2005multi}, where VPBO makes a selection on the categorical inputs $\bm{h}_t$ using the value proposal set $\mathcal{V}_t$. The maximum acquisition value for a proposal out of the proposal set $\mathcal{V}_t$ is selected and its corresponding input $[\bm{h}^*_t, \bm{x}^*_t]$ is chosen as the next query point of the objective function $f$. Following this, the newly observed function value $f^*_t$ and corresponding input $z^*_t$ is added to the observation set $\mathcal{D}_t$.

The result is that when selecting the value proposal, we are able to choose from all the usual acquisition metrics (\eg UCB, EI, \etc) and use this single metric to select both the categorical \textit{and} continuous components of the input. The surrogate GP model is the singular driving mechanism behind our entire method, all whilst factoring in the uncertainty of the surrogate in the decision making process for the discrete and continuous sub-spaces of the input. This is in contrast to other hybrid MAB-BO frameworks which require separate standalone bandits in addition to a BO back-end~\cite{nguyen2020bayesian, ru2020bayesian}.

In summary, our method is described as follows: at each iteration $t$, for each possible combination of categorical choices, we employ the GP-based BO back-end and optimise the continuous sub-space for each combination of categories, given our categorical variables. The resulting optimisation over the continuous sub-space produces a value proposal for each combination of categories and we make a selection on the decision variables in the discrete sub-space which maximises the proposal value over all possible categorical combinations. Our VPBO procedure is detailed in Algorithm~\ref{alg:vpbo_pseudo}.

\subsection{Theoretical Analysis}
\label{ssec:theoretical_analysis}

We now show that our method converges to the global maximum with a sub-linear rate in the mixed space setting. At optimisation iteration $t$, VPBO performs the following:
\begin{align}
    \notag [\bm{h}^*_t, \bm{x}^*_t] &= \argmax_{\bm{z}} \alpha_t(\bm{z}|\mathcal{D}_{t-1}) \\
    &= \argmax_{\bm{h}} \argmax_{\bm{x}} \alpha_t(\bm{x}|\mathcal{D}_{t-1}, \bm{h}).
\label{eq:vpbo_opt}
\end{align}

\begin{algorithm}[h]
\caption{VPBO Optimisation}
\label{alg:vpbo_pseudo}
\textbf{Input} Black-box function $f$, Initial observation data $\mathcal{D}_0$, Maximum number of iterations $T$, Number of possible categorical combinations $C$ \\
\textbf{Output} The best recommendation $\bm{z}^*_T = [\bm{h}^*_T , \bm{x}^*_T]$
\begin{spacing}{0.8}
\begin{algorithmic}[1]
\State Initialise the data $\mathcal{D}_0$
\ForAll {$t = \{1,...,T\}$}
    \State Fit GP using $\mathcal{D}_{t-1}$
    \ForEach {$c = \{1,...,C\}$}
    \State $\bm{x}^*_{t,c} = \underset{\bm{x}}{\argmax{}} \alpha_{t,c}(\bm{x} | \mathcal{D}_{t-1}, \bm{h}_{t,c})$
    \State $v_{t,c} = \alpha_{t,c}(\bm{x}^*_{t,c} | \mathcal{D}_{t-1}, \bm{h}_{t,c})$
    \State $\mathcal{V}_t[c] = [\bm{x}^*_{t,c}, v_{t,c}]$   \Comment{Add $[\bm{x}^*_{t,c}, v_{t,c}]$ to proposal set}
    \EndFor
    \State $c^*$ = $\underset{c}{\argmax{}} \mathcal{V}_t$
    \State Set $\bm{z}^*_t = [\bm{h}^*_t, \bm{x}^*_t]$ = $[\bm{h}_{t,c^*}, \bm{x}^*_{t,c^*}]$
    \State Query at $\bm{z}^*_t$ to obtain $f^*_t$
    \State Augment the data: $\mathcal{D}_t \leftarrow \mathcal{D}_{t-1} \bigcup (\bm{z}^*_t, f^*_t)$
\EndFor
\end{algorithmic}
\end{spacing}
\end{algorithm}

This is equivalent to a Bayesian Optimisation setting over a mixed search space, where each continuous component of the mixed search space corresponds to a combination of categories $c$. We consider the noisy case $y_{c,t} = f_c(\bm{x}) + \xi_t$, where $\xi_t \sim \mathcal{N}(0, \sigma_\xi^2)$. To simplify our notation, when referencing in the context of Eq.~\ref{eq:vpbo_opt}, we drop the $c$ term from $\bm{x}_c$ and $f_c$, using the notation $\bm{x} \in \mathcal{X}_c$ to resolve any ambiguity. Following~\cite{srinivas2009gaussian}, we assume our objective $f$ is smooth according to the Reproducing Kernel Hilbert Space (RKHS) associated with a GP with a mixed-kernel $\kappa_z$ in the mixed search space setting:
\begin{asu}
The objective function $f$ is a member of the Reproducing Kernel Hilbert Space $\mathcal{H}_{\kappa_z}([\mathcal{C}, \mathcal{X}])$ with known kernel function $\kappa_z$, and has a bounded norm $||f||_{\mathcal{H}_{\kappa_z}} \leq B$, for some $B > 0$.
\label{asu:mixed_space_rkhs}
\end{asu}
This assumption is central for Bayesian Optimisation in justifying the use of GPs to estimate $f$ from samples~\cite{rasmussen2003gaussian} and ensuring our corresponding regret bounds hold. To show the convergence of VPBO in the mixed search space setting, we assume $f$ is bounded in $[\mathcal{C}, \mathcal{X}]$, which is common as it is generally assumed in BO that $f$ is Lipschitz continuous~\cite{brochu2010tutorial}:
\begin{asu}
The objective function $f(\bm{z})$ is bounded in $[\mathcal{C}, \mathcal{X}]$, \ie $\exists F_1, F_2 \in \mathbb{R} : \forall \bm{z} \in [\mathcal{C}, \mathcal{X}], F_1 \leq f(\bm{z}) \leq F_2$.
\label{asu:lipschitz_bounded}
\end{asu}

Using Assumptions~\ref{asu:mixed_space_rkhs} and~\ref{asu:lipschitz_bounded} and the assumptions from~\cite{vazquez2010convergence}, we can show the optimisation in Eq.~\ref{eq:vpbo_opt} can be treated as a general BO problem, with the surrogate defined as a  GP with mixed-kernel $\kappa_z$ and using Expected Improvement as our acquisition criterion $\alpha$; under this setting, $\kappa_z$ should be a valid kernel~\cite{bull2011convergence}.
\begin{lemma}
The mixed kernel $\kappa_z$ is a valid kernel (\ie positive semi-definite).
\begin{proof}
The categorical kernel $\kappa_h$ is demonstrated to be a valid kernel in~\cite{ru2020bayesian}. The mixed kernel $\kappa_z$ is additive and multiplicative between two valid kernels $\kappa_h$ and $\kappa_x$ which results in a valid kernel. Therefore, the mixed kernel $\kappa_z$ is also a valid kernel. This concludes the proof.
\end{proof}
\label{lem:valid_mixed_kernel}
\end{lemma}
We relax the continuous kernel assumption of~\cite{vazquez2010convergence} and derive the maximum information gain of our mixed-kernel $\kappa_z$. We define the \textit{maximum information gain} complexity measure~\cite{srinivas2009gaussian} as:
\begin{definition}
\label{def:max_info_gain}
Given a sequence of choices $A = \{\bm{z}_1,...,\bm{z}_T\} \subset \mathcal{Z}$, let $f_A = \{f(\bm{z}_i)\}$, $y_A = \{f(\bm{z}_i) + \xi_i\}$ and $I$ be the mutual information. The maximum information gain $\psi_T$ after $T$ iterations is defined as:
\begin{equation}
\psi_T := \max_{A \subset \mathcal{Z}, |A|=T} I(y_A;f_A).
\end{equation}
\end{definition}

We show that the information gain of the mixed-kernel after $T$ iterations, $\psi_T^z$, is bounded by some maximum value. For convenience, we denote the maximum information gain on the mixed kernel as $\psi_T^z := \psi_T(\kappa_z; [\mathcal{C}, \mathcal{X}])$, that on the categorical kernel as $\psi_T^h := \psi_T(\kappa_h; \mathcal{C})$ and that on the continuous kernel as $\psi_T^x := \psi_T(\kappa_x; \mathcal{X})$.

\begin{lemma}
For the mixed-kernel $\kappa_z$, the maximum information gain $\psi_T^z$ after $T$ iterations is given by:
\begin{align}
    \notag \psi_T^z &\leq \mathcal{O}\bigl((\lambda C + 1 -\lambda)(T^{1-2\eta}(\log T)) \\
    & + (C + 2 - 2\lambda)\log T \bigr),
    \label{eq:mixed_kernel_info_gain_final}
\end{align}
where $\eta = \frac{\nu}{2\nu + d_x(d_x + 1)}$, $\nu > 2$ and $d_x \geq 1$ are the smoothness parameter and dimension of the \matern{} kernel $\kappa_x$, $C$ is the number of possible categorical combinations and $\lambda \in [0,1]$ is the trade-off parameter between the sum and product kernels of $\kappa_h$ and $\kappa_x$.
\begin{proof}
From~\cite{krause2011contextual}, given two kernels $\kappa_h$ and $\kappa_x$, and if $\kappa_h$ is a kernel on $\mathcal{C}$ with rank at most $m$, then the bound on the maximum information gain for the sum and product of these kernels is respectively given as: 
\begin{align}
    \psi_T(\kappa_h + \kappa_x; [\mathcal{C}, \mathcal{X}]) \leq \psi_T^h + \psi_T^x + 2\log T \label{eq:sum_max_info_gain} \\
    \psi_T(\kappa_h \kappa_x; [\mathcal{C}, \mathcal{X}]) \leq m\psi_T^x + m\log T \label{eq:prod_max_info_gain}.
\end{align}

Using~\cite{zhou2011simple, kirchhoff2020gaussian}, we can find the maximum rank $r$ of $\kappa_h$ by applying a Cholesky-type decomposition such that $r = \prod_{j=1}^{k} N_j = C$, where $k$ is the number of categorical variables each with $N_j$ distinct values. Applying this to Eq.~\ref{eq:sum_max_info_gain} and~\ref{eq:prod_max_info_gain}, we have bounds on the information gain of the sum and product kernels for $\kappa_h$ and $\kappa_x$ respectively:
\begin{align}
    \psi_T(\kappa_h + \kappa_x; [\mathcal{C}, \mathcal{X}]) \leq \mathcal{O}(\psi_T^x + (C+2)\log T) \label{eq:sum_max_info_gain_2} \\
    \psi_T(\kappa_h \kappa_x; [\mathcal{C}, \mathcal{X}]) \leq C \psi_T^x + C \log T \label{eq:prod_max_info_gain_2}.
\end{align}
The mixed-kernel $\kappa_z$ is defined as $\kappa_z = \lambda \kappa_h\kappa_x + (1-\lambda)(\kappa_h+\kappa_x)$ where $\lambda \in [0,1]$~\cite{ru2020bayesian}. Using Eq.~\ref{eq:sum_max_info_gain_2} and~\ref{eq:prod_max_info_gain_2} with this definition, we obtain the following bound on the information gain of the mixed-kernel $\kappa_z$:
\begin{align}
    \notag \psi_T^z &\leq \lambda\mathcal{O}(C\psi_T^x + C\log T) \\ 
    \notag &+ (1-\lambda)(\psi_T^x + (C+2)\log T) \\ 
    &\leq \mathcal{O}((\lambda C + 1 -\lambda)\psi_T^x + (C + 2 - 2\lambda)\log T).
    \label{eq:mixed_kernel_info_gain_1}
\end{align}

From~\cite{srinivas2009gaussian}, the maximum information gain on the continuous component of the mixed-kernel $\psi_T^x$, is given as: $\psi^x_T := \psi_T(\kappa_x;\mathcal{X}) \sim \mathcal{O}(T^{1-2\eta}(\log T))$, where $\eta = \frac{\nu}{2\nu + d_x(d_x + 1)}$. Applying this to Eq.~\ref{eq:mixed_kernel_info_gain_1} yields Eq.~\ref{eq:mixed_kernel_info_gain_final}, thus concluding the proof.
\end{proof}
\label{lem:mixed_kernel_info_bound}
\end{lemma}

We now bound the regret at iteration $T$, $\mathcal{R}_T$, for our method:
\begin{theorem}
\label{lem:conv_rate_for_ei_mixed_kernel}
Let $\sigma^2_{\xi}$ be the measurement noise variance, $\beta_T = 2||f||^2_{\kappa_z} + 300\psi^z_T \log^3(T/\delta)$ and $\delta \in (0,1)$. Then, using $y^{max} = \underset{y_i \in \mathcal{D}_t}{\max} y_i$ as the incumbent, the cumulative regret of EI with mixed-kernel $\kappa_z$ is given by the sub-linear rate:
\begin{equation}
    \mathcal{R}_T \leq \sqrt{T\beta_T\psi^z_T} \sim \mathcal{O} \left(\sqrt{T^{1-\eta}(\log T)^{3}} \right),
\label{eq:conv_rate_for_ei_mixed_kernel}
\end{equation}
with probability at least $1-\delta$.

\begin{proof}
Under the valid kernel setting of Lem.~\ref{lem:valid_mixed_kernel} and applying Lem.~\ref{lem:mixed_kernel_info_bound} to Theorem 4 from~\cite{nguyen2017regret}, we obtain Eq.~\ref{eq:conv_rate_for_ei_mixed_kernel} and thus conclude the proof.
\end{proof}
\end{theorem}
Therefore, the regret for our method increases at a sub-linear rate $\mathcal{R}_T \sim \mathcal{O}\left(\sqrt{T^{1-\eta}(\log T)^{3}} \right)$, which vanishes in the limit as $\lim_{T \rightarrow \infty} \frac{\mathcal{R}_T}{T} = 0$.

\section{Experiments}
\label{sec:results}

\begin{figure*}[h]
\centering
\subfloat[Func2C]{
\includegraphics[width=0.24\textwidth]{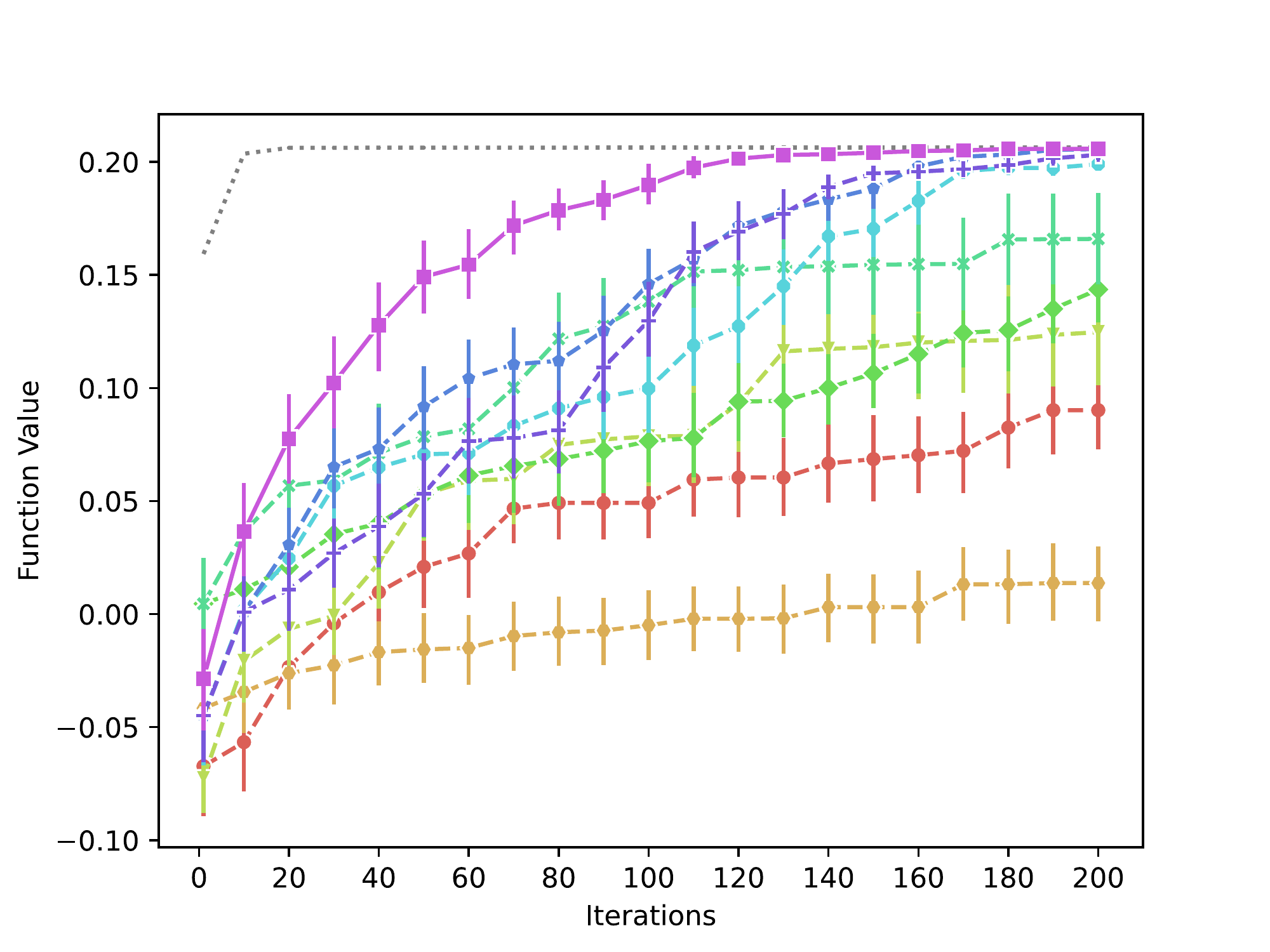}
\label{fig:func2c_bestvals}
}
\subfloat[Func3C]{
\includegraphics[width=0.24\textwidth]{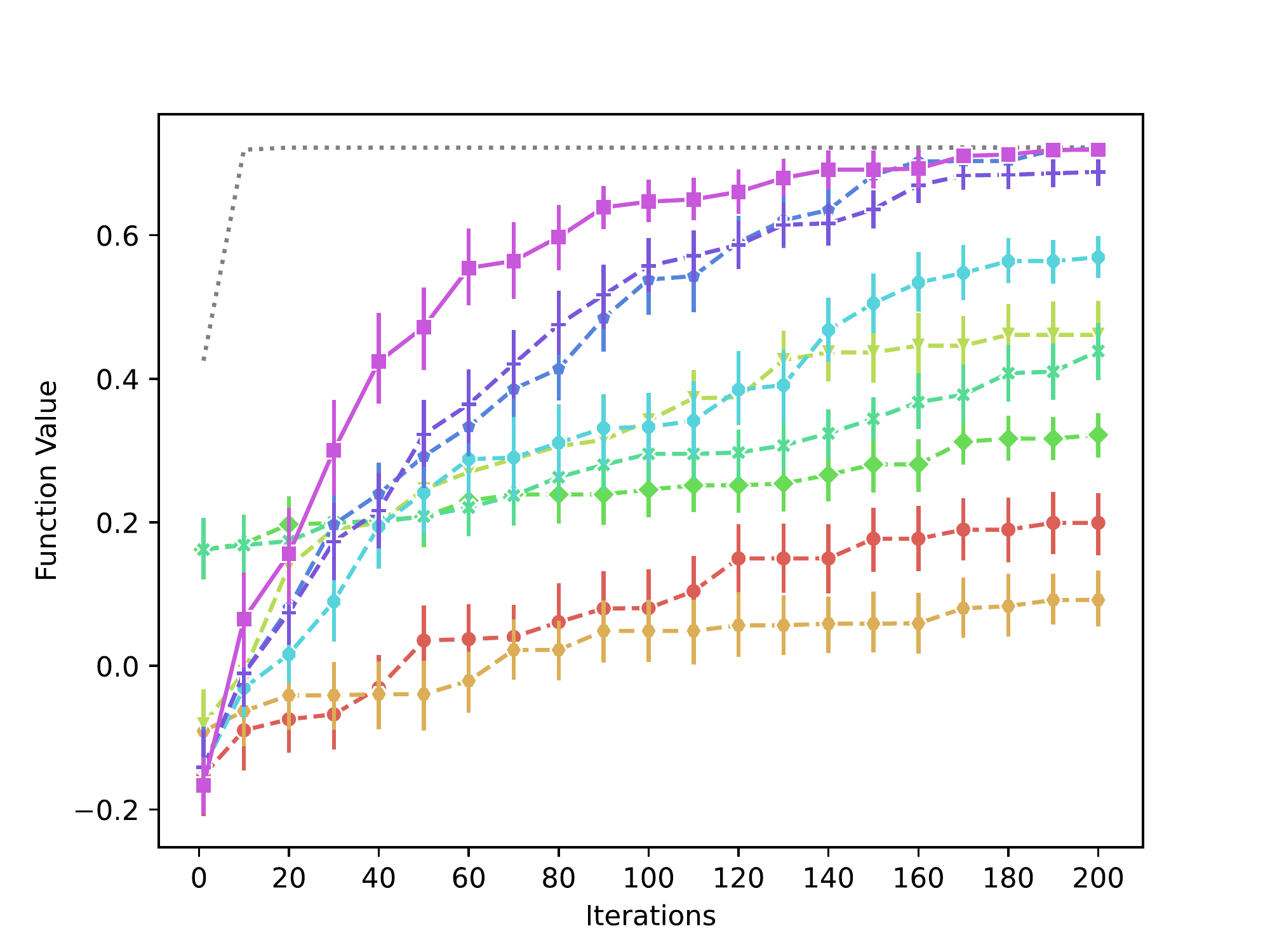}
\label{fig:func3c_bestvals}
}
\subfloat[Reizman-Suzuki]{
\includegraphics[width=0.24\textwidth]{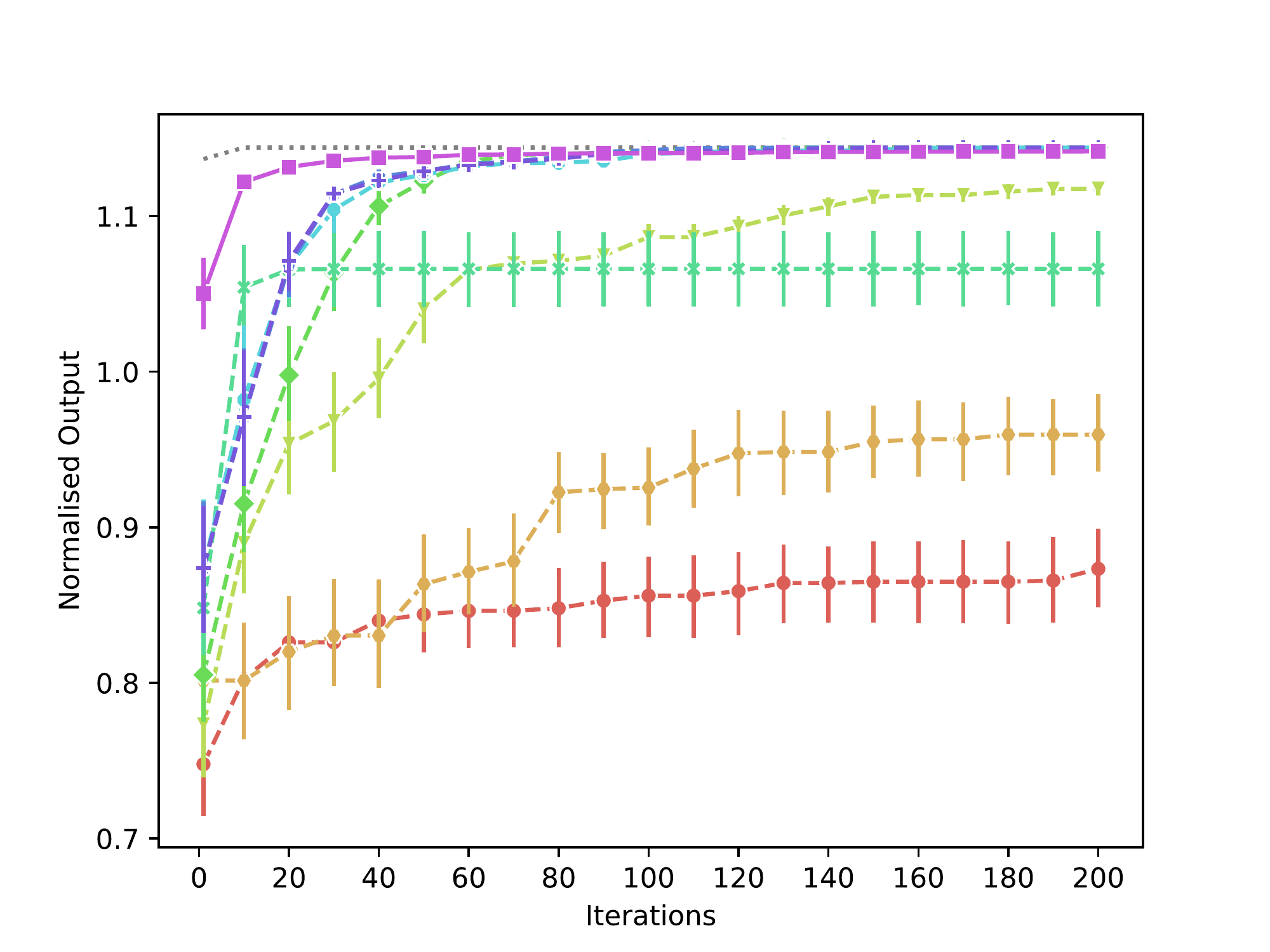}
\label{fig:suzuki_bestvals}
}
\subfloat[Baumgartner]{
\includegraphics[width=0.24\textwidth]{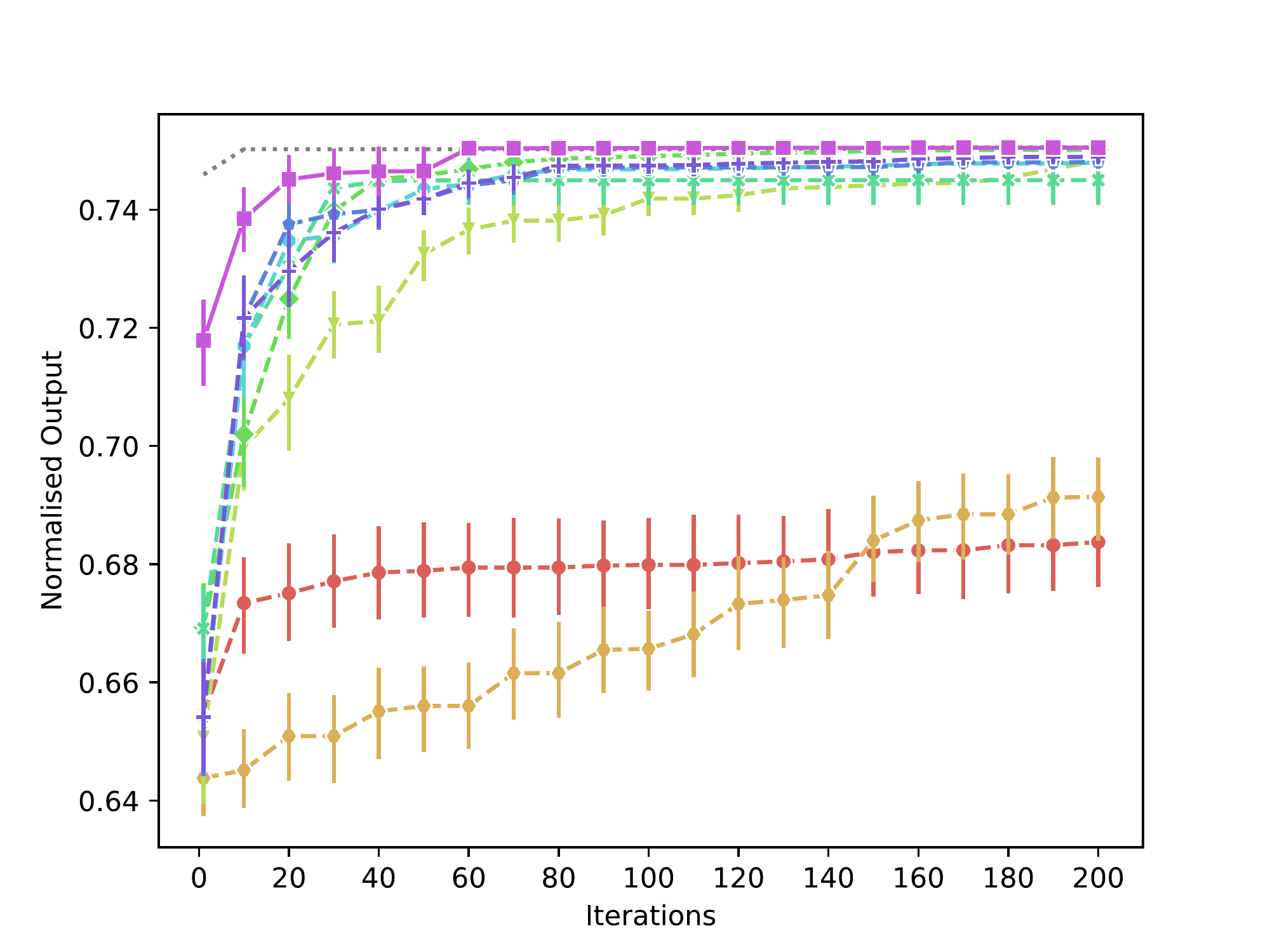}
\label{fig:baumgartner_bestvals}
} \\
\vspace{-1.0em}
\subfloat[SVM-Boston]{
\includegraphics[width=0.32\textwidth]{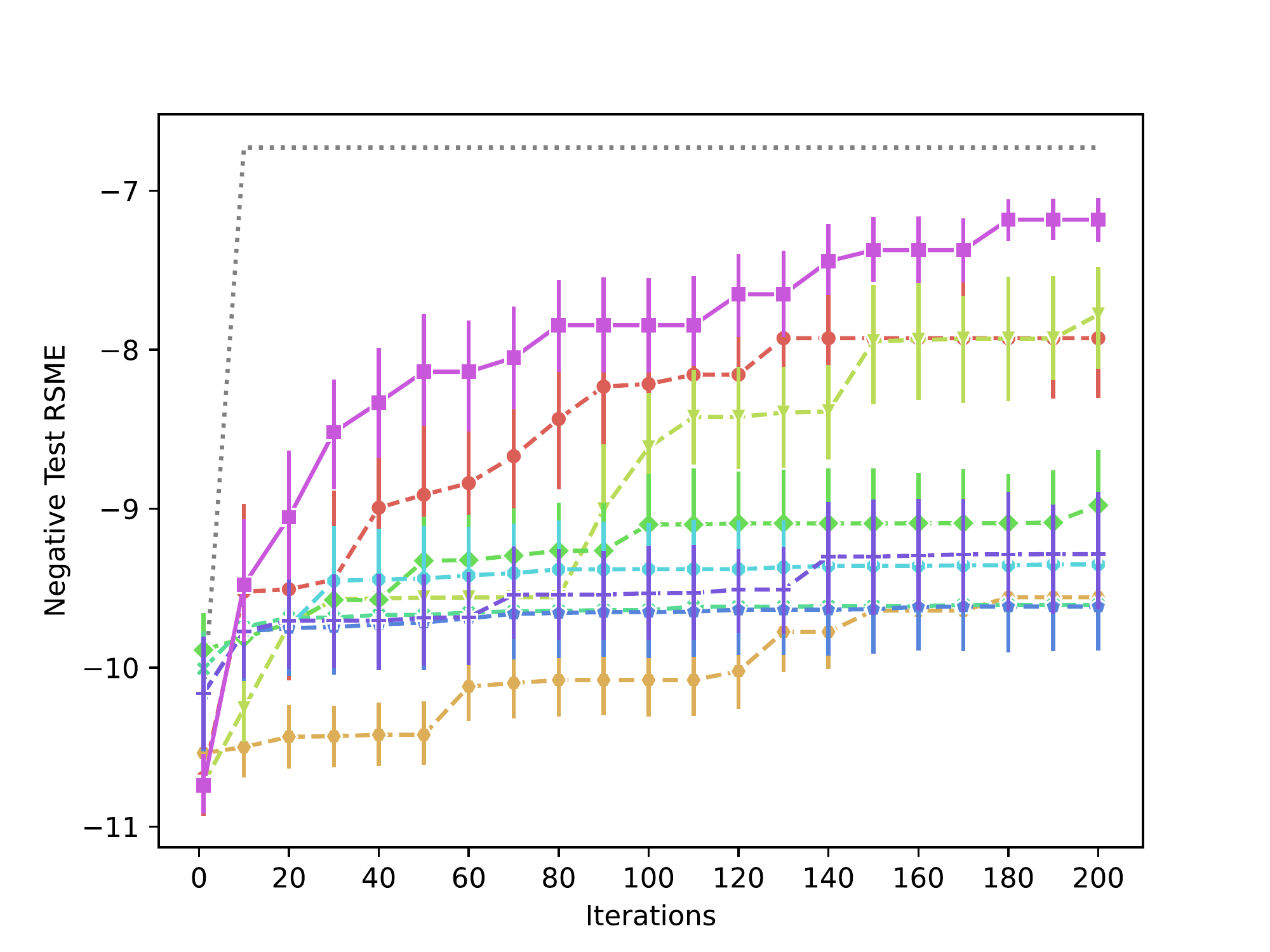}
\label{fig:svmboston_bestvals}
}
\subfloat[XG-MNIST]{
\includegraphics[width=0.32\textwidth]{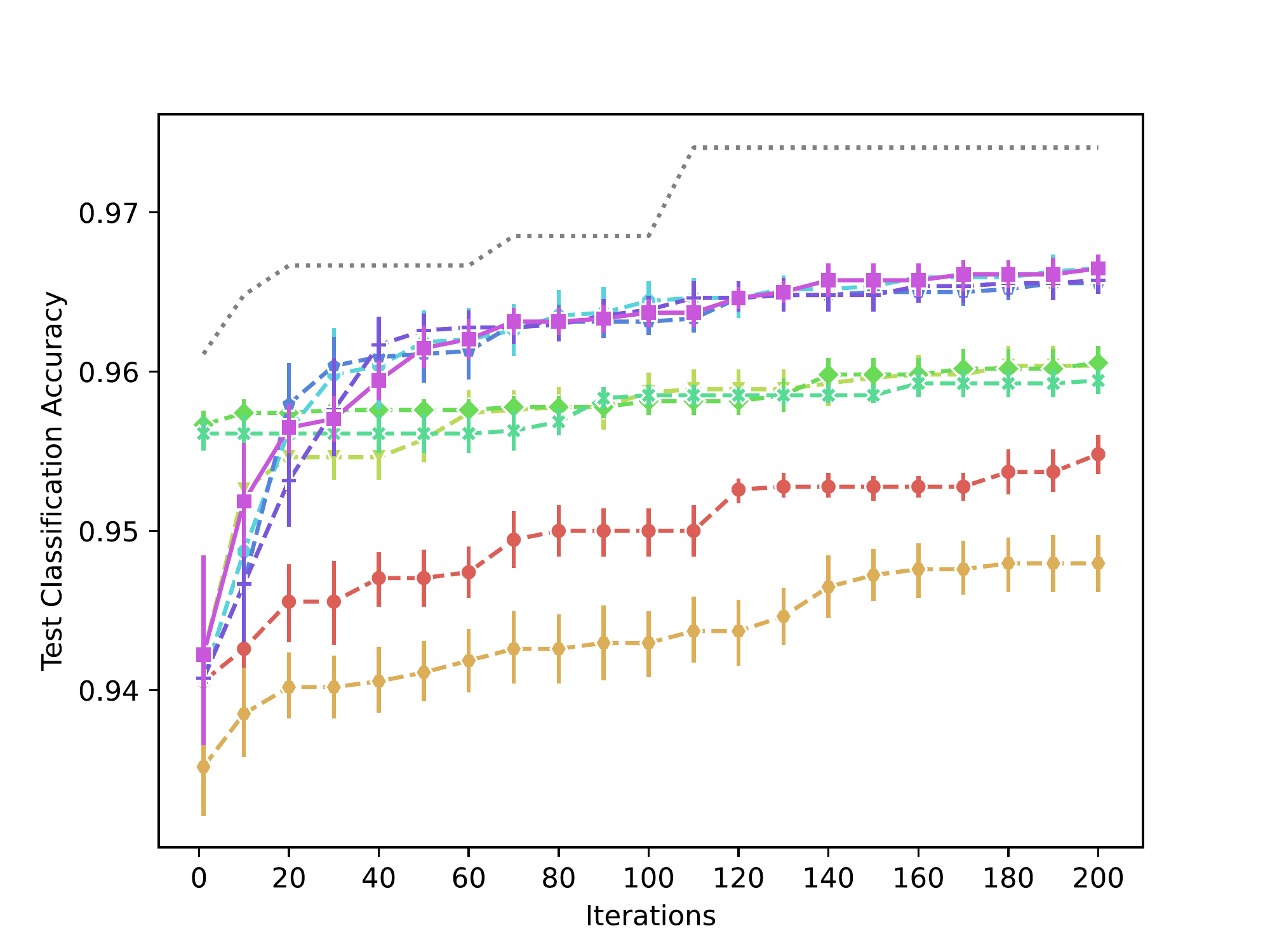}
\label{fig:xgmnist_bestvals}
}
\subfloat[NASBench-101]{
\includegraphics[width=0.32\textwidth]{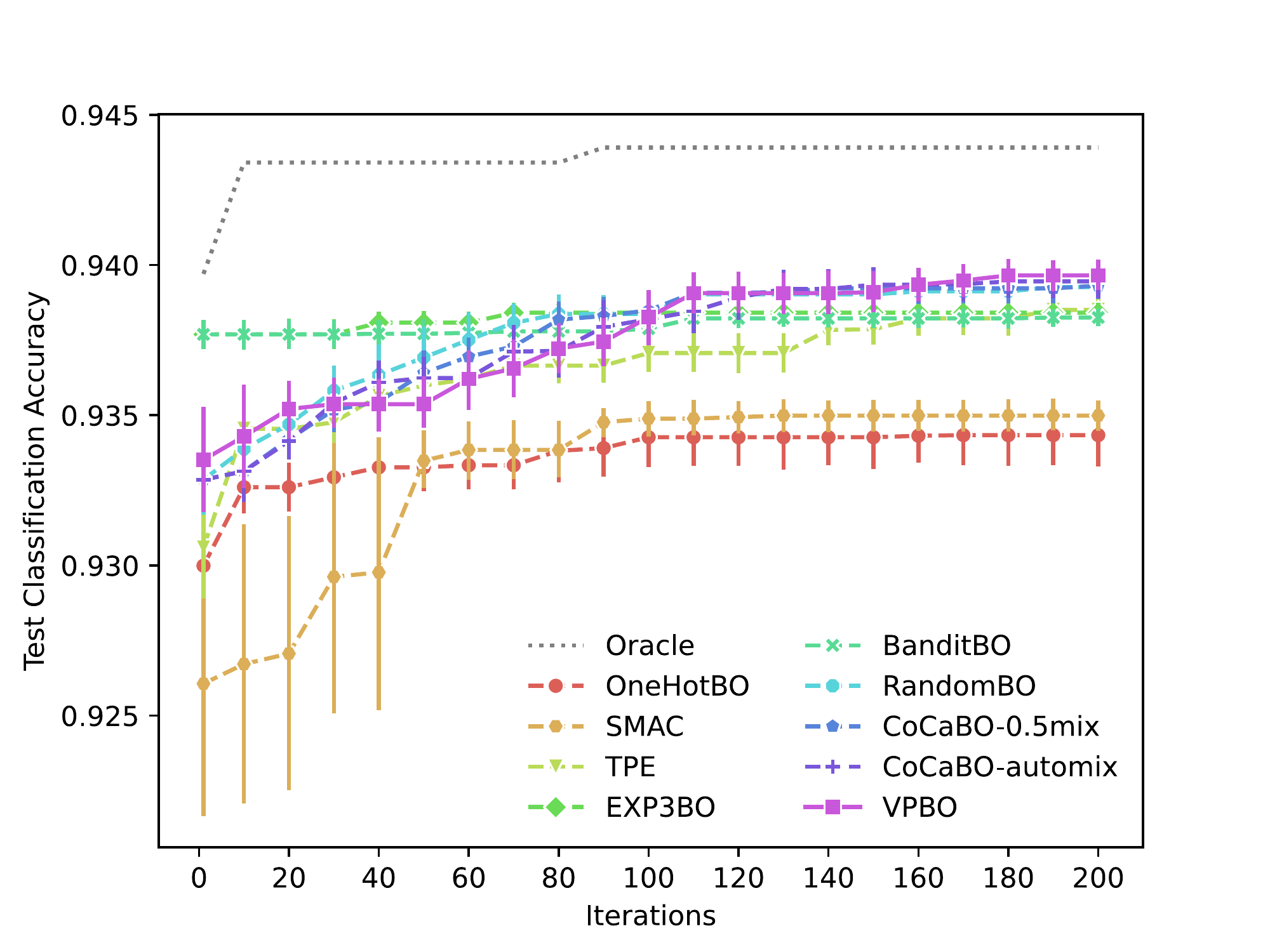}
\label{fig:nasbench101_bestvals}
}
\caption{Performance of VPBO against existing methods on various synthetic and real-world tasks.}
\label{fig:bestvals}
\end{figure*}

We compared VPBO against several competing baselines which can handle mixed-variable type inputs: SMAC~\cite{hutter2011sequential}, TPE~\cite{bergstra2011algorithms}, GP-based BO with one-hot encoding (One-Hot BO), EXP3BO~\cite{gopakumar2018algorithmic}, Bandit-BO~\cite{nguyen2020bayesian} and CoCaBO~\cite{ru2020bayesian}. All baseline methods are implemented according to their publicly available Python implementations\footnote{One-Hot BO: \url{https://github.com/SheffieldML/GPyOpt}, SMAC: \url{https://github.com/automl/SMAC3}, TPE: \url{https://github.com/hyperopt/hyperopt}, EXP3BO: \url{https://github.com/shivapratap/AlgorithmicAssurance\_NIPS2018}, Bandit-BO: \url{https://github.com/nphdang/Bandit-BO}, CoCaBO: \url{https://github.com/rubinxin/CoCaBO\_code}}. Additionally, we implemented RandomBO, where the EXP3 bandit algorithm in CoCaBO is replaced by a bandit that selects categorical variables at random. 

\paragraph{Competing Baseline Settings}
One-Hot BO, EXP3BO, Bandit-BO, RandomBO and CoCaBO use the Upper Confidence Bound (UCB) acquisition function~\cite{srinivas2009gaussian} with trade-off parameter $k=2.0$. For CoCaBO, we used the two best performing variations reported in the paper, using kernel mix values of $\lambda=\{\texttt{0.5}, \texttt{auto}\}$, where $\lambda=\texttt{auto}$ indicates $\lambda$ being optimised as a hyperparameter. All experiments are run under the sequential setting (\ie batch size of 1). 

For SMAC, TPE, RandomBO and CoCaBO, we initialise each model with $24$ randomly sampled initial points as was done in~\cite{ru2020bayesian}. Both EXP3BO and Bandit-BO require more than $24$ initial samples since these approaches need to fit a separate surrogate model for each categorical combination. That is, the initial observation data needs to be evenly divided and allocated to surrogate models for each categorical combination. We follow the approach of~\cite{ru2020bayesian} and initialise each surrogate model with $3$ randomly sampled initial points. 

\paragraph{VPBO Settings}
For our VPBO method, we utilise the same mixed-kernel GP surrogate as~\cite{ru2020bayesian}, replacing the UCB acquisition function with the Expected Improvement (EI) acquisition function~\cite{mockus1978application, jones1998efficient}. 

In order to make VPBO computationally tractable for datasets with large $C$, we relax the optimisation on the continuous component of the input, uniformly sampling $200$ samples for each BO iteration (compared to $5000$ samples required for the CoCaBO baseline). As shown in the results, $200$ samples was sufficient in practice for VPBO to achieve convergence and by relaxing the optimisation on the continuous component, we are able to achieve improved mean wall-clock time over CoCaBO (refer to Appendix~\ref{app:wallclock_overhead}). We follow the optimisation approach of~\cite{ru2020bayesian}, optimising GP hyperparameters by maximising the log marginal likelihood every 10 iterations using multi-started gradient descent, including the mixed-kernel hyperparameter (\ie $\lambda=\texttt{auto}$). 

For our initialisation budget, we also allocate an initial $24$ observation samples. Since our method relies on the surrogate for making choices on both the categorical and continuous parts of the input, it is important that the GP model is initialised correctly. In particular, a good choice of initial observation data can significantly improve the performance of our VPBO method. To ensure an appropriate choice of initial observations is made for our surrogate GP model, we employ a search initialisation procedure, where we use the BO back-end to search for $\bm{x}^*_0$. The search initialisation process is described as follows: we first randomly sample up to half the initial observation budget (\ie $12$ points) as usual, randomly selecting $\bm{h}_0^j$ and $\bm{x}_0^j$ where $j \in \{1,...,12\}$. We then continue to randomly select $\bm{h}_0^j$, whilst using the BO back-end to search for $\bm{x}_0^j$ for the remaining initial observation budget (\ie $j \in \{13,..24\})$. 

To search for $\bm{x}_0^j$, we use the Max-Value Entropy Search (MES) acquisition metric of~\cite{wang2017max}, which has been shown to be efficient at optimising output space information gain and fulfils the criteria we are interested in during the initial observation stage. In Sec.~\ref{ssec:searchinit_ablation}, we perform a detailed ablation study on this search initialisation method, comparing our method to the CoCaBO baseline.

\begin{figure}[h]
\centering
\includegraphics[width=0.45\textwidth]{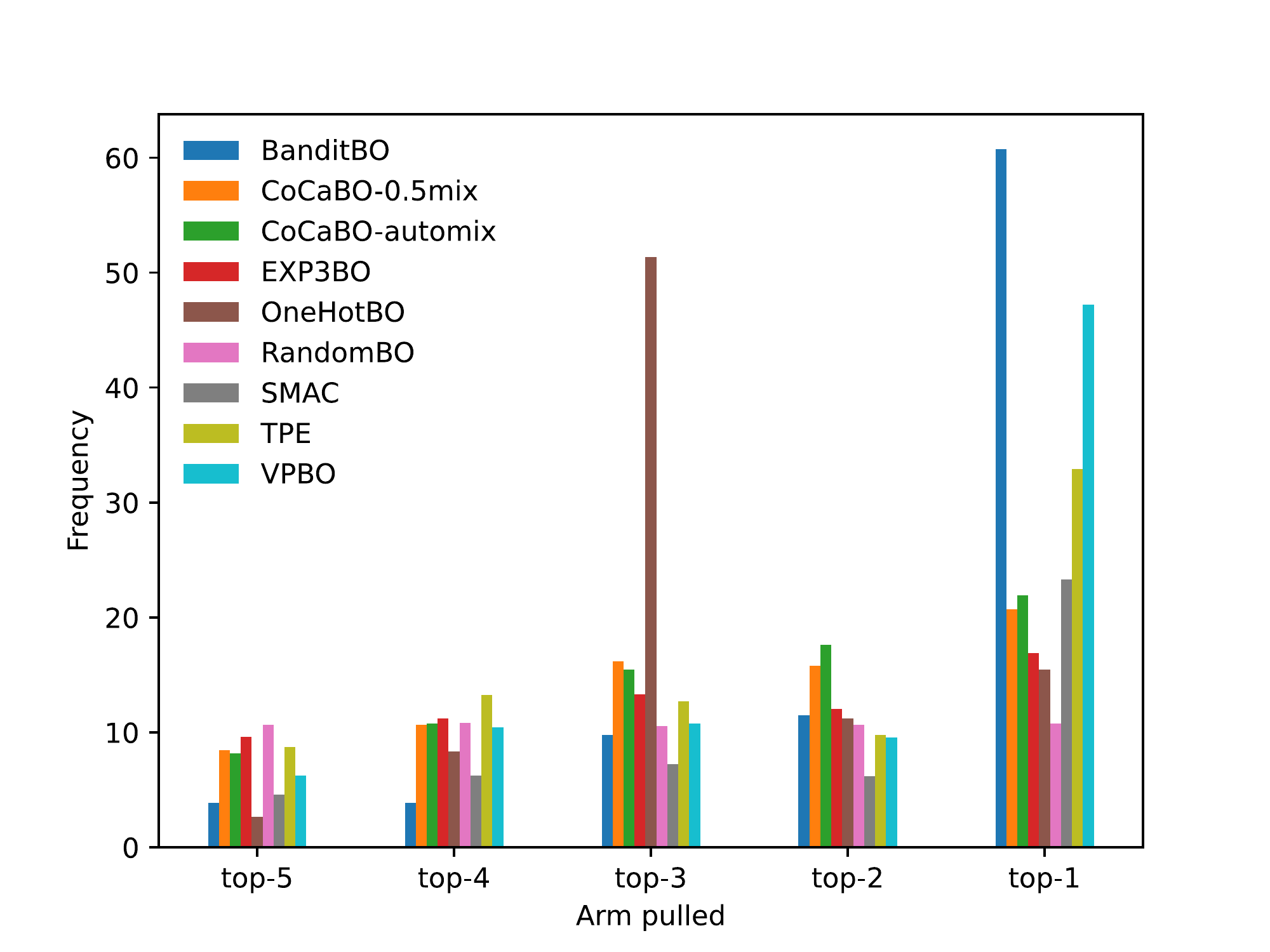}
\caption{Top 5 average best arm pull frequency of all methods across all 7 datasets.}
\label{fig:top5_arm_pulls}
\end{figure} 

\paragraph{Datasets}
For all benchmarks, the continuous inputs were normalised to $\bm{x} \in [0,1]^{d_x}$ and all experiments were conducted on an 8-core 3.4Ghz Intel Xeon processor with 64GB RAM. Our benchmarks include a variety of synthetic and real, single and multi-objective problems:

\begin{itemize}[leftmargin=*]
    \item \textbf{Func-2C} is a synthetic test problem with two continuous variables ($d_x=2$) and two categorical variables ($k=2$). The categorical variables determine the linear combination of three $2D$ global optimisation benchmark functions: Beale, Six-Hump Camel and Rosenbrock\footnote{\url{https://www.sfu.ca/~ssurjano/optimization.html}} (15 possible combinations).
    \item \textbf{Func-3C} is an extension of \textbf{Func-2C} with an additional categorical variable ($k=3)$, allowing for more possible categorical choices (60 possible combinations).
    \item \textbf{Reizman-Suzuki} represents virtual experiments for the Suzuki-Miyaura Cross-Coupling reaction where experimental outcomes are based on an emulator that is trained on the experimental data published by~\cite{reizman2016suzuki}. The experimental emulator is provided by~\cite{felton2020summit} and outputs product yield and catalyst turnover number as objectives to be maximised. This optimisation problem consists of three continuous variables ($d_x=3$): temperature, residence time and catalyst loading; and one categorical variable ($k=1$): catalyst choice (8 possible combinations).
    \item \textbf{Baumgartner} represents virtual experiments for the Aniline Cross-Coupling reaction where experimental outcomes are based on an emulator that is trained on the experimental data published by~\cite{baumgartner2019use}. The experimental emulator is provided by~\cite{felton2020summit} and outputs product yield and material cost as objectives to be maximised and minimised respectively.  This optimisation problem consists of three continuous variables ($d_x=3$): temperature, residence time and base equivalents; and two categorical variables ($k=2$): catalyst and base choices (12 possible combinations).
    \item \textbf{SVM-Boston} outputs the test negative mean square error from a support vector machine (SVM) for regression on the Boston Housing dataset~\cite{dua2017uci}. This optimisation problem consists of three continuous variables ($d_x=3$): nu value, error term penalty and tolerance; and three categorical variables ($k=3$): kernel type, kernel coefficient and shrinking (16 possible combinations).
    \item \textbf{XG-MNIST} outputs the test classification accuracy of an XGBoost model~\cite{chen2015xgboost} on MNIST~\cite{lecun1998mnist}. This optimisation problem consists of four continuous variables ($d_x=4$): learning rate, minimum split loss, subsampling ratio and regularisation; with four categorical variables ($k=4$): booster type, growth policy, training objective and maximum tree depth (80 possible combinations).
    \item \textbf{NAS-Bench-101} outputs the test classification accuracy from performing architecture search on the convolutional neural network topology for CIFAR-10 image classification~\cite{krizhevsky2014cifar}. The search was conducted using the NAS-Bench-101 dataset using the proposed search space from~\cite{ying2019bench}. This optimisation problem consists of $22$ continuous variables ($d_x=22$) corresponding to edge probabilities and five categorical variables ($k=5$) of intermediate node types (243 possible combinations).
\end{itemize}

\begin{figure*}[h]
\centering
\subfloat[Func2C]{
\includegraphics[width=0.24\textwidth]{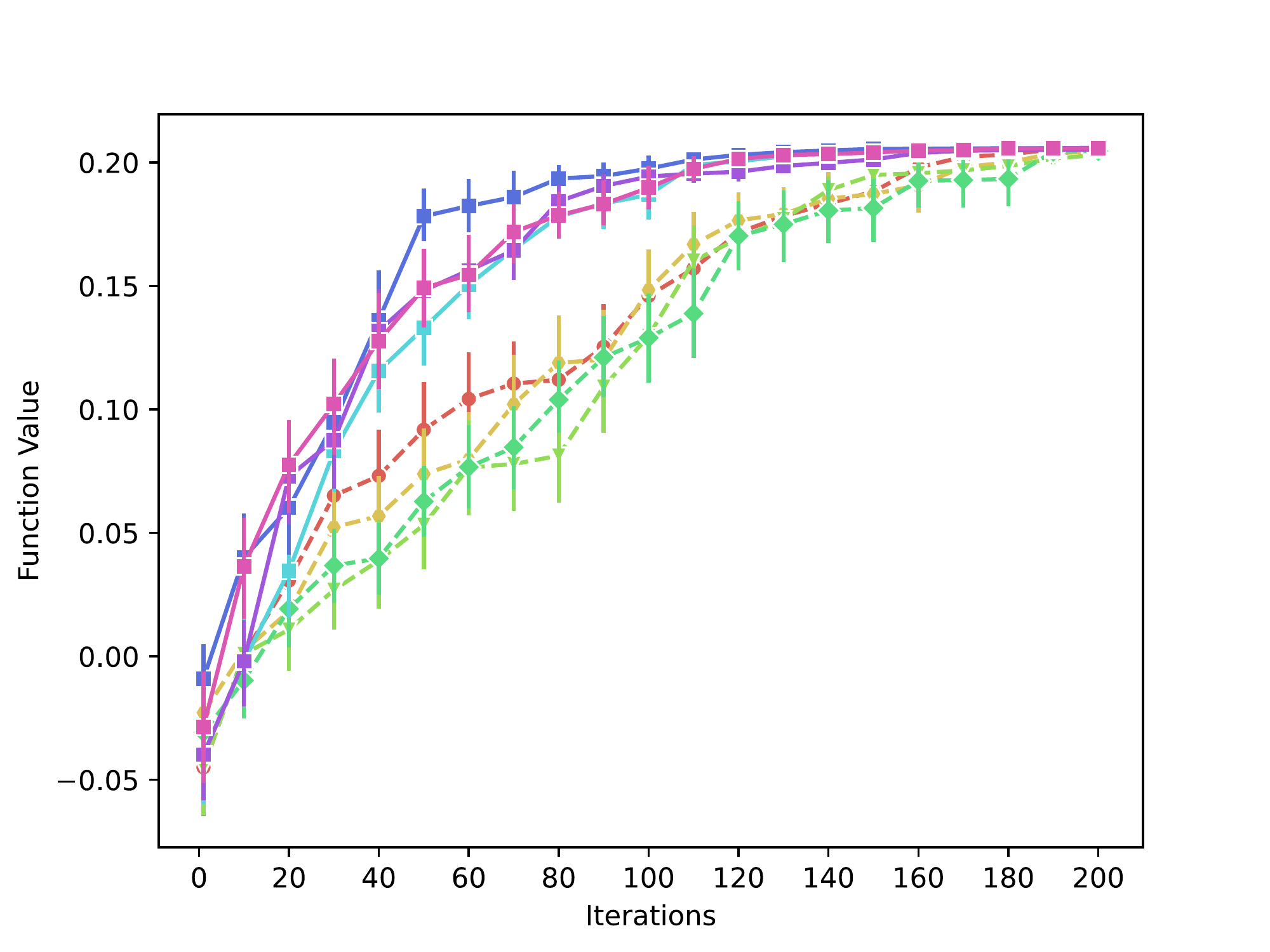}
\label{fig:func2c_init_ablation}
}
\subfloat[Func3C]{
\includegraphics[width=0.24\textwidth]{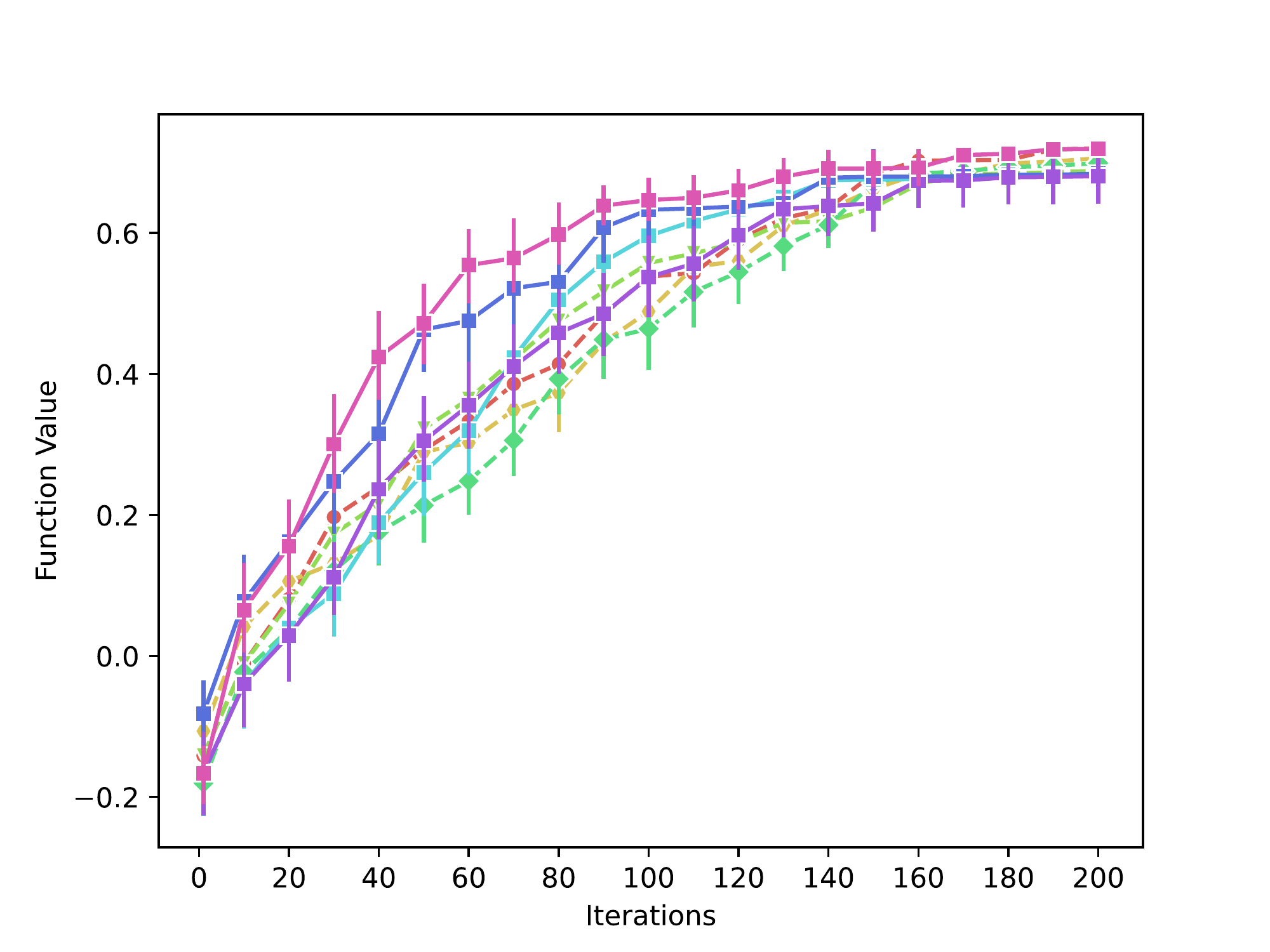}
\label{fig:func3c_init_ablation}
}
\subfloat[Reizman-Suzuki]{
\includegraphics[width=0.24\textwidth]{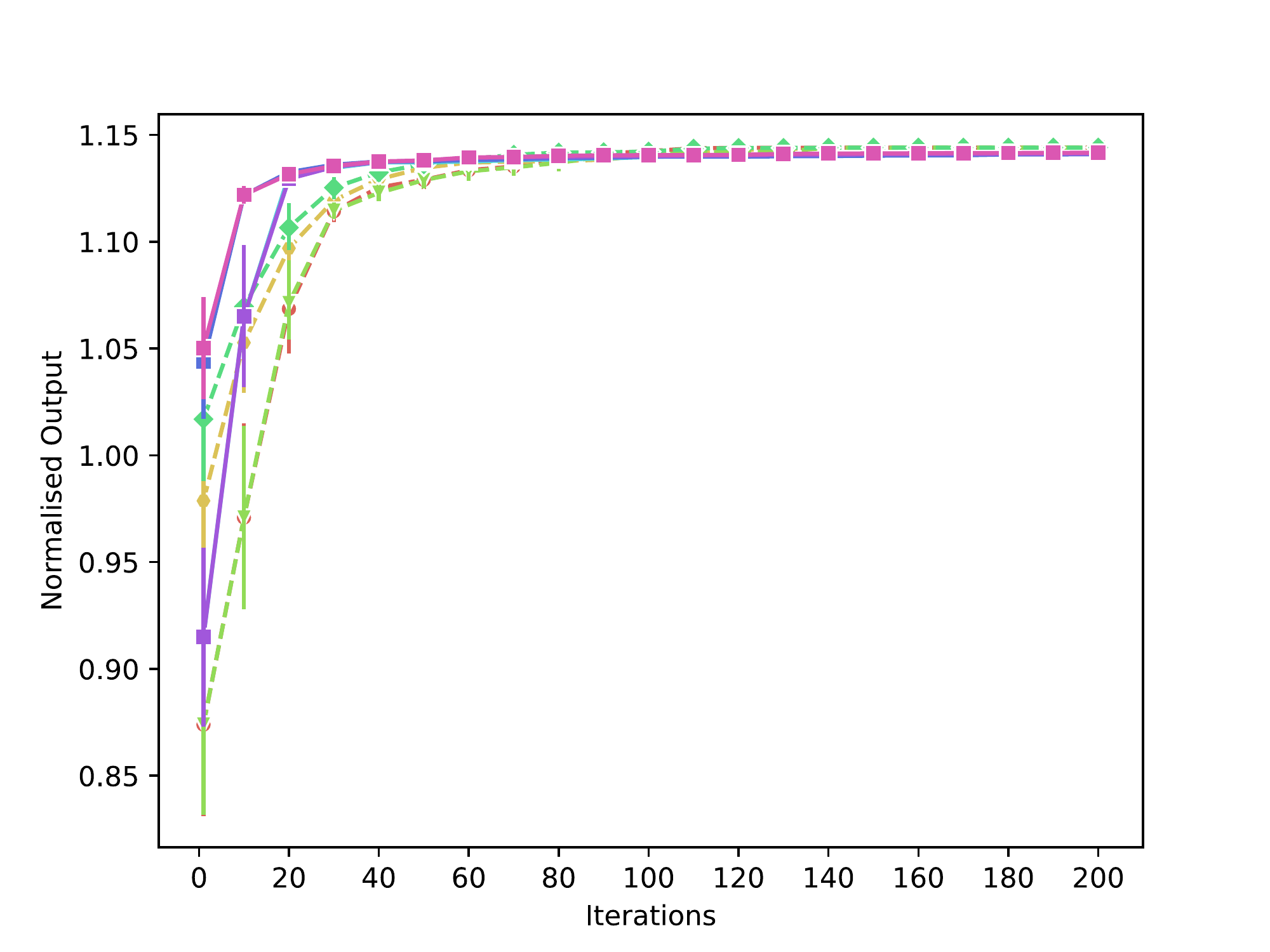}
\label{fig:suzuki_init_ablation}
}
\subfloat[Baumgartner]{
\includegraphics[width=0.24\textwidth]{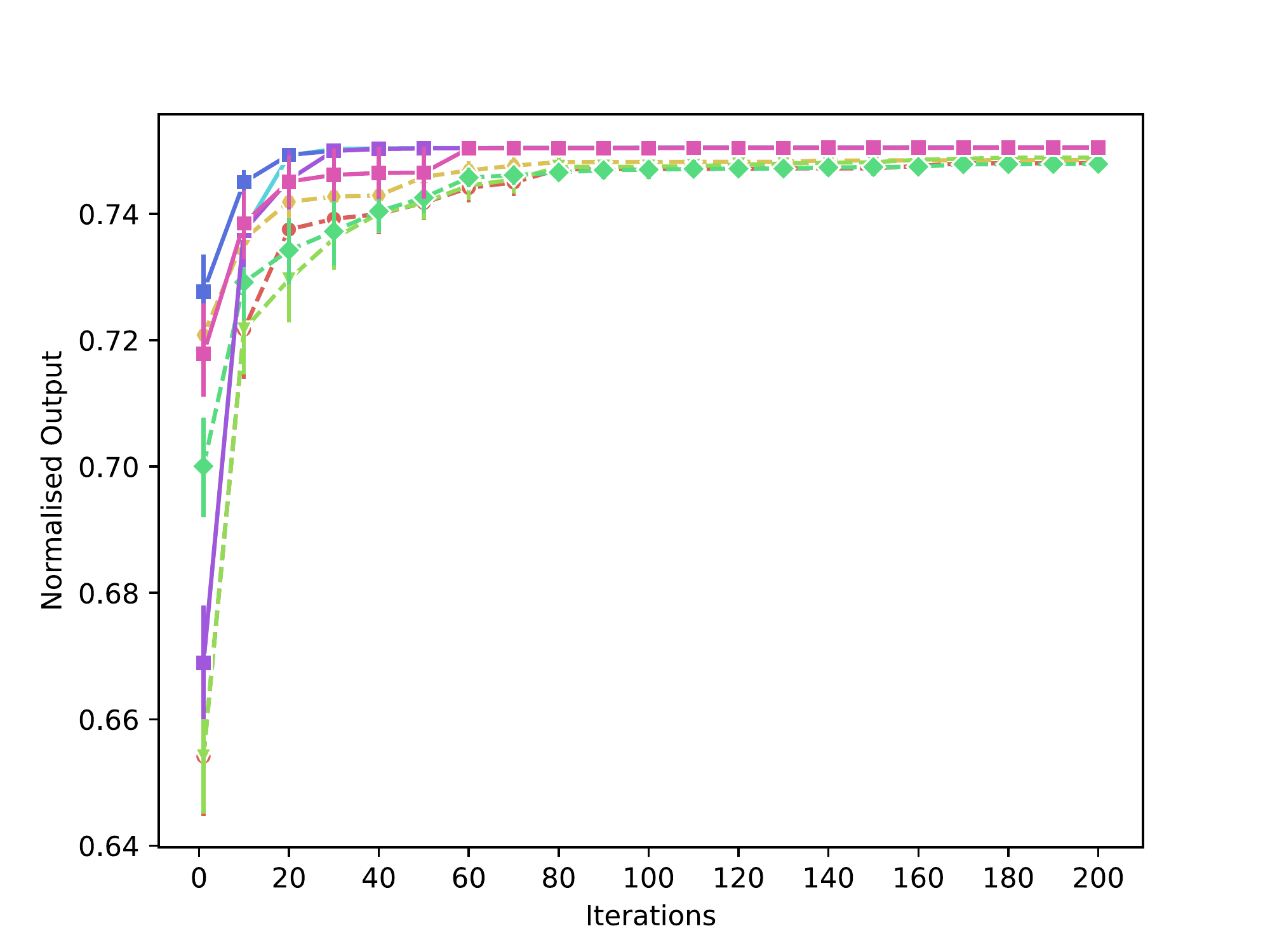}
\label{fig:baumgartner_init_ablation}
} \\
\vspace{-1.0em}
\subfloat[SVM-Boston]{
\includegraphics[width=0.32\textwidth]{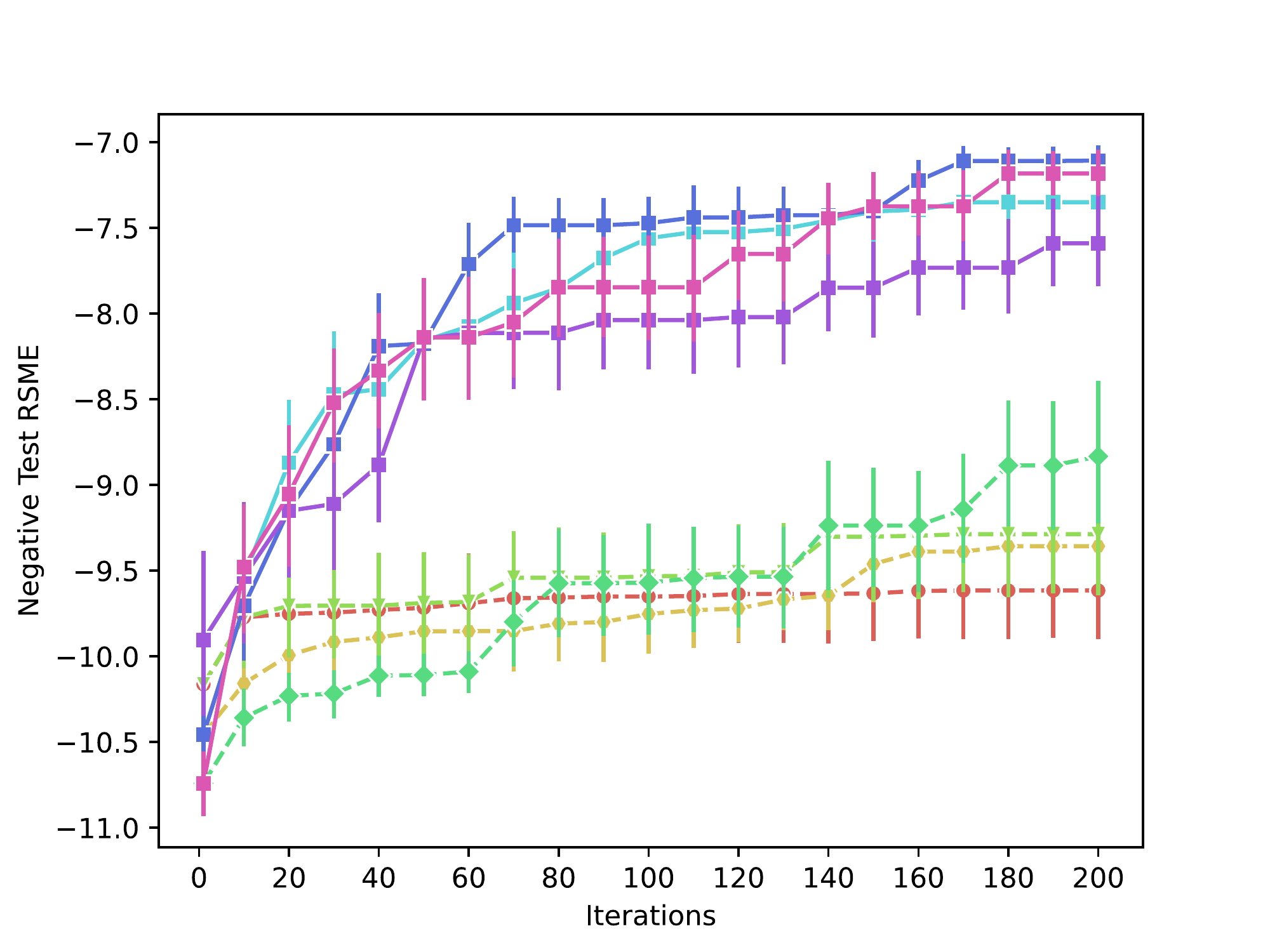}
\label{fig:svmboston_init_ablation}
}
\subfloat[XG-MNIST]{
\includegraphics[width=0.32\textwidth]{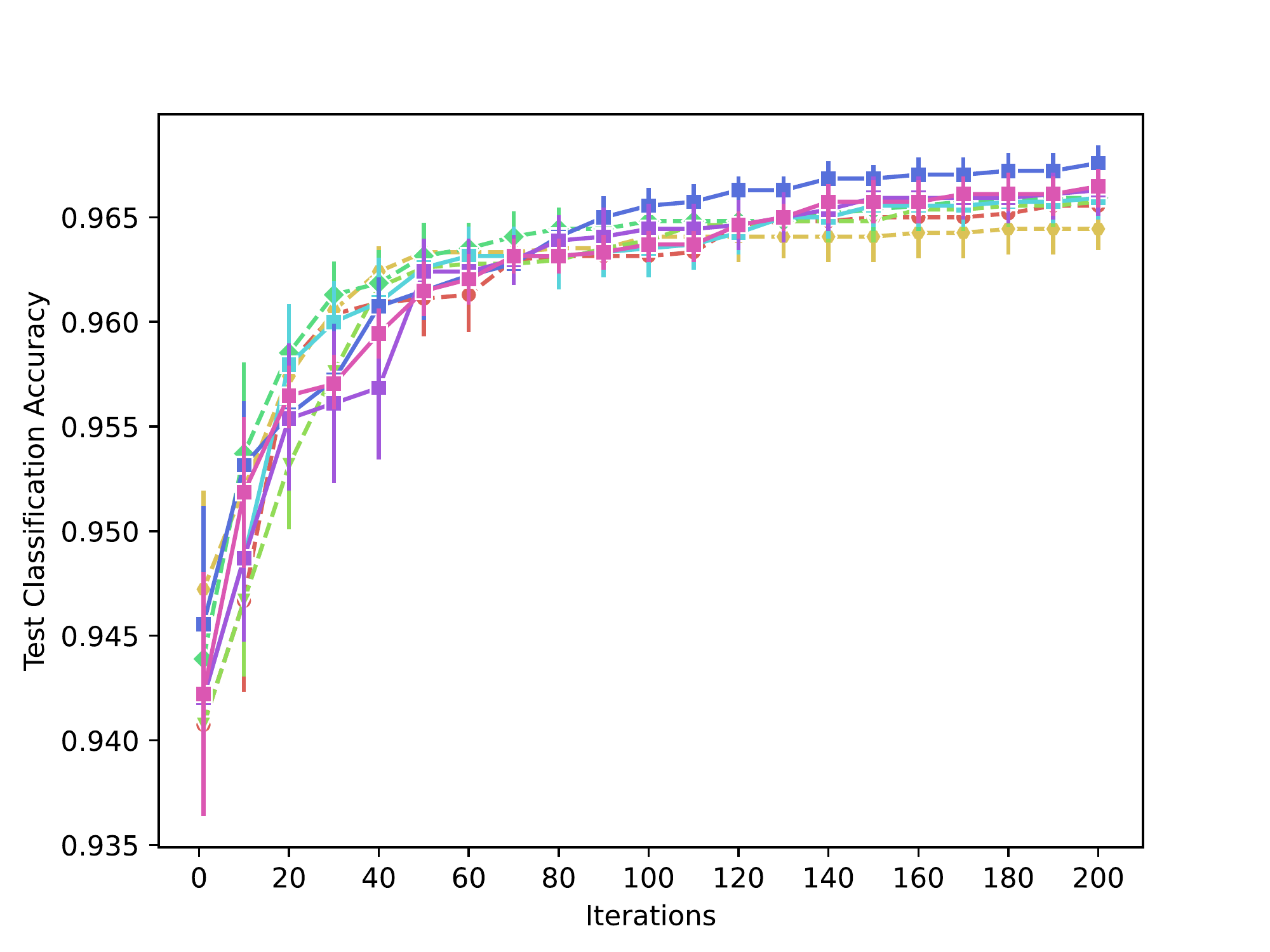}
\label{fig:xgmnist_init_ablation}
}
\subfloat[NASBench-101]{
\includegraphics[width=0.32\textwidth]{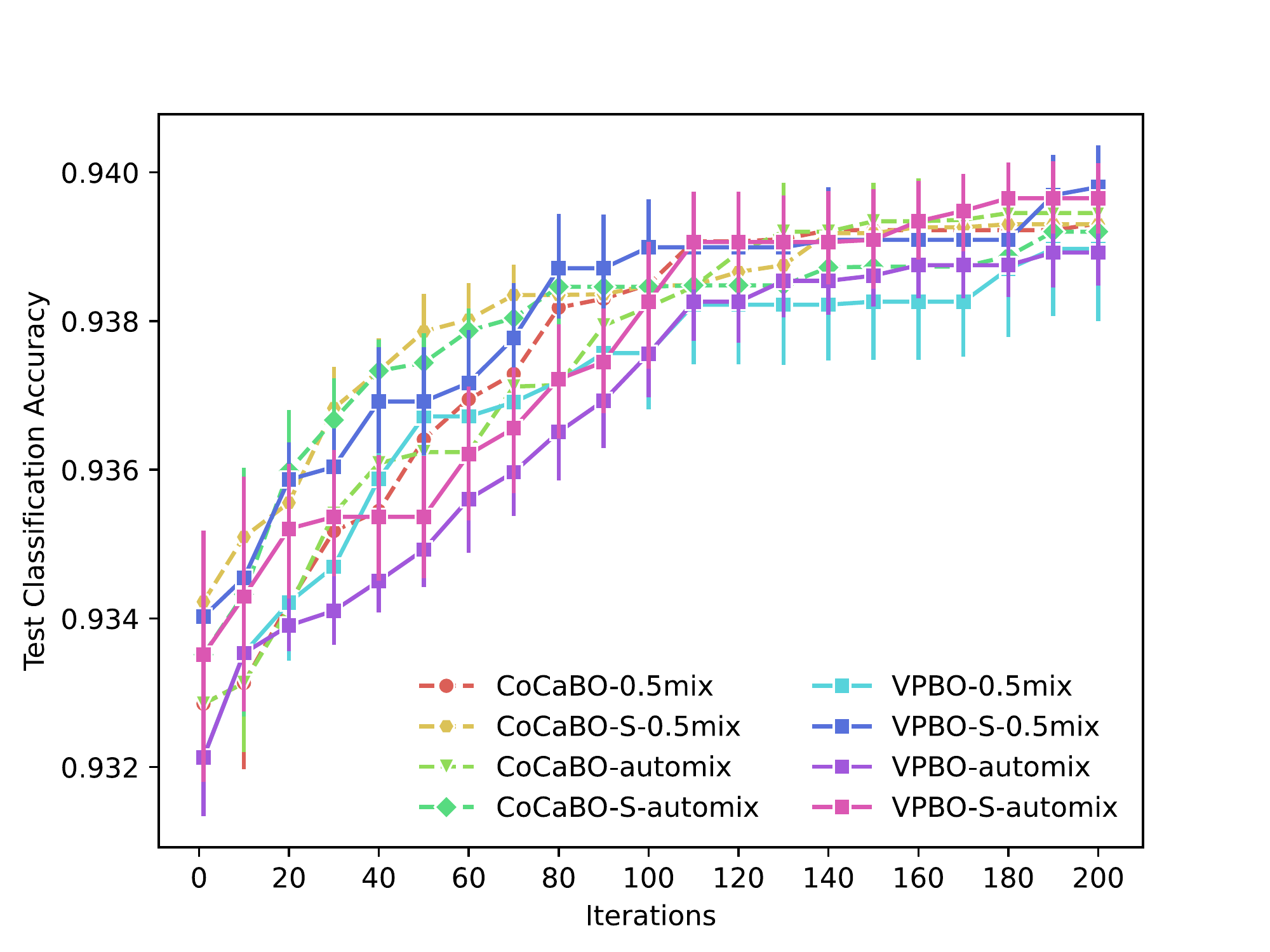}
\label{fig:nasbench101_init_ablation}
}
\caption{Ablation study on search initialisation procedure between CoCaBO and our proposed VPBO method}
\label{fig:searchinit_ablation}
\end{figure*} 

\subsection{Performance of VPBO}
\label{ssec:vpbo_performance}
We evaluated the optimisation performance of our proposed VPBO method against existing methods. Following~\cite{ru2020bayesian}, we set each optimisation trial to consist of $T=200$ iterations. We performed $20$ random trials for the Func2C, Func3C, Reizman-Suzuki and Baumgartner datasets. For the SVM-Boston, XG-MNIST and NASBench-101 datasets, we performed 10 random trials. Means and standard errors over all trials are presented in Fig.~\ref{fig:bestvals}.

The gray dotted line in Fig.~\ref{fig:bestvals} represents an Oracle agent that we trained (similar to~\cite{gopakumar2018algorithmic}). Here, we run BO for each of the $C$ choices of possible category combinations, with each choice allocated its own separate GP surrogate model $\text{GP}_c$. We allocate each of the $C$ choices the full $T=200$ iterations as well as all $24$ initial sample points and optimise the hyperparameters for $\text{GP}_c$ every iteration by maximising the log marginal likelihood. At each iteration $t$, the Oracle's performance is then taken as the best value for $f$ over all possible categorical choices $C$.  Hence, we use the Oracle to represent a best possible outcome scenario at each iteration $t$.

Across all synthetic and real-world problems, our VPBO method outperforms other competing approaches, where the general trend is that VPBO demonstrates a significant improvement in initial performance and more quickly converges towards the performance of the Oracle agent when compared with the competing baselines. Note that due to the nature of the EXP3BO and Bandit-BO methods maintaining separate surrogate models for each of the possible $C$ categories and being allocated $3$ initial samples per surrogate, for problems where $C$ is large (\eg Func3C, XG-MNIST and NASBench-101), their initial performance at $t=$1 is significantly higher due to the larger number of initial samples observed from $f$ (\eg $180$ samples for Func3C vs. $24$ for all other methods including VPBO).

\subsection{Choice of Categorical Variables}
We compare our VPBO method against existing baselines in terms of how often they make the best selection of categorical variables. Using bandit problem terminology, we term each combination of categories an arm and selecting a combination of categories which corresponds to a particular arm is referred to as ``pulling" that arm. The best selection is defined as the set of choices found to yield the best results according to the Oracle agent described in Sec.~\ref{ssec:vpbo_performance}. Ideally, our agent should choose the best performing arm often, so that the continuous component of the input is given many opportunities to find the optimal $\bm{x}^*_t$ at any given iteration $t$. Fig.~\ref{fig:top5_arm_pulls} shows the average arm pull frequency of VPBO and all baselines on the top 5 performing arms as indicated by the Oracle, averaged across the 7 datasets. We can see that VPBO outperforms all baselines except Bandit-BO, where it offers competitive results. Note that due to the nature of the Bandit-BO algorithm, it must be allocated significantly more initial observation samples (as detailed in Sec.~\ref{sec:results}) before an optimisation trial, when compared to VPBO.

\subsection{Ablation on Search Initialisation}
\label{ssec:searchinit_ablation}
We perform an ablation study of the effect of our search initialisation strategy when compared to the typical fully random initialisation approach. The results across the $7$ datasets are shown in Fig.~\ref{fig:searchinit_ablation}. Here, we compare our method to the baseline of~\cite{ru2020bayesian}, selecting the two best performing variants, \texttt{0.5mix} and \texttt{automix}. The former fixes $\lambda=0.5$, while the latter learns $\lambda$ as part of the GP hyperparameters. CoCaBO-S and VPBO-S indicate where search initialisation has been employed. Using search initialisation significantly improves the performance of our VPBO method but in general makes a negligible difference in CoCaBO. As shown in Fig.~\ref{fig:searchinit_ablation}, our VPBO variants tend to outperform CoCaBO variants regardless of the initialisation procedure.


\section{Conclusion}
\label{sec:conclusion}
In this paper, we presented a mixed-variable black-box optimisation approach which adopts a global BO approach in the joint optimisation of categorical and continuous variables. Our solution uses a single acquisition criterion for optimising the mixed input input space under a unified framework. We analyse our method theoretically and empirically, demonstrating that VPBO has a sub-linear bound on regret and shows significant improvements in performance over existing mixed-variable optimisation approaches on a wide range of problems. Future work would be to extend our model into additional search spaces such as graphs and into higher dimensional categorical settings.

\newpage
\bibliographystyle{plain}
\bibliography{jmlr-sample}

\newpage
\appendix

\section{Derivation of the Closed-form for Expected Improvement}
\label{app:EI_closed_form}
Let $\mathcal{D}_T = \{x_t \in \mathbb{R}^d, y_t \in \mathbb{R}\}^T_{t=1}$ be the set of observations up to time $T$. The improvement function at iteration $t$ is given by:
\begin{equation}
    \mathcal{I}_t(\bm{x}_t) = \max{} \{0, f(\bm{x}_t) - \mathcal{E}_t\}
\label{eq:improvement_function_2}
\end{equation}
where $\mathcal{E}_t$ defines an incumbent at iteration $t$, such that $\mathcal{E}_t = y_{t-1}^{max} = \underset{y_i \in \mathcal{D}_{t-1}}{\max{}} y_i$. For concise notation, we drop the $\bm{x}$ term and denote $\mathcal{I}$ as the improvement function, $\mu$ as the predictive mean of the posterior and $\sigma$ as the predictive variance of the posterier. The likelihood of improvement $\mathcal{I}$ on a normal posterior distribution is given as:
\begin{equation}
    Pr(\mathcal{I}) = \frac{1}{\sqrt{2\pi}\sigma}\exp{\left(-\frac{(\mu - \mathcal{E} - \mathcal{I})^2}{2\sigma^2}\right)}
\label{eq:improvement_likelihood}
\end{equation}

The expected improvement is defined as $\alpha^{EI}(\bm{x}) = \mathbb{E}[\mathcal{I}(\bm{x})]$. Using Eq.~\ref{eq:improvement_likelihood}, we can rewrite the expected improvement as:
\begin{equation}
    \alpha^{EI}(\bm{x}) = \int_{0}^{\infty} \frac{\mathcal{I}}{\sqrt{2\pi}\sigma} \exp{\left(-\frac{(\mu - \mathcal{E} - \mathcal{I})^2}{2\sigma^2}\right)} d\mathcal{I}
\end{equation}

Let $t=\frac{\mu - \mathcal{E} - \mathcal{I}}{\sigma}$, then $\mathcal{I} = -t\sigma + \mu - \mathcal{E}$ and $dt = -\frac{1}{\sigma}d\mathcal{I}$. We can rewrite $\alpha^{EI}(\bm{x})$ as:

{\tiny
\begin{align}
    \notag \alpha^{EI}(\bm{x}) = \int_{t=\frac{\mu - \mathcal{E}}{\sigma}}^{-\infty} \frac{-t\sigma + \mu - \mathcal{E}}{\sqrt{2\pi}\sigma} \exp{\left(-\frac{t^2}{2}\right)}(-\sigma) dt \\
    = \int_{t=\frac{\mu - \mathcal{E}}{\sigma}}^{-\infty} \frac{t}{\sqrt{2\pi}} \exp{\left(-\frac{t^2}{2}\right)} dt + (\mu - \mathcal{E})\int^{t=\frac{\mu - \mathcal{E}}{\sigma}}_{-\infty} \frac{1}{\sqrt{2\pi}}\exp{\left(-\frac{t^2}{2}\right)}dt
\label{eq:improvement_likelihood_2}
\end{align}
}

Denoting $u = t^2 = \left(\frac{\mu - \mathcal{E} - \mathcal{I}}{\sigma} \right)^2$, $du=2tdt$, and the first term in Eq.~\ref{eq:improvement_likelihood_2} is as follows:

{\tiny
\begin{align}
    \notag \int_{t=\frac{\mu - \mathcal{E}}{\sigma}}^{-\infty} \frac{t}{\sqrt{2\pi}} \exp{\left(-\frac{t^2}{2}\right)} dt = \frac{\sigma}{\sqrt{2\pi}} \int_{u=t^2}^{-\infty} \exp{\left(-\frac{u}{2}\right)} \frac{du}{2} \\
    \notag = \frac{\sigma}{\sqrt{2\pi}} \left[ -\exp{\left(-\frac{1}{2} \left(\frac{\mu - \mathcal{E} - \mathcal{I}}{\sigma}\right)^2\right)} \right]^{\mathcal{I}=-\infty}_{\mathcal{I}=0} \\
    = \sigma\mathcal{N}\left( \frac{\mu-\mathcal{E}}{\sigma} | 0,1 \right)
\label{eq:improvement_likelihood_first_term}
\end{align}
}

The second term in Eq.~\ref{eq:improvement_likelihood_2} is computed as:
\begin{align}
    \notag (\mu - \mathcal{E})\int^{t=\frac{\mu - \mathcal{E}}{\sigma}}_{-\infty}\frac{1}{\sqrt{2\pi}}\exp{\left(-\frac{t^2}{2}\right)}dt \\
    \notag =  (\mu - \mathcal{E})\int^{0}_{-\infty} \frac{1}{\sqrt{2\pi}}\exp{\left(-\frac{1}{2}\left(\frac{\mu - \mathcal{E} - \mathcal{I}}{\sigma}\right)^2\right)}d\gamma \\
    = (\mu - \mathcal{E}) \Phi\left(\frac{\mu-\mathcal{E}}{\sigma}\right)
\end{align}

Denoting $\gamma = \frac{\mu-\mathcal{E}}{\sigma}$, we obtain the closed-form for the EI acquisition function:
\begin{equation}
    \alpha^{EI}(\bm{x}) = \sigma(\bm{x})\phi(\gamma) + (\mu(\bm{x}) - \mathcal{E})\Phi(\gamma)
\end{equation}
where $\phi(\gamma) = \mathcal{N}(\gamma |0,1)$ and $\Phi(\gamma)$ are the respective standard normal PDF and CDF.


\section{Mixed Kernel}
\label{app:mixed_kernel}
\paragraph{Categorical Kernel}
For our categorical kernel, we use an indicator function  based on what was proposed in~\cite{xu2009multi} ($1_\mathbb{C}$), where $1_\mathbb{C}$ is an indicator function which equals $1$ when its condition $\mathbb{C}$ (in this case $h=h'$) is met and $0$ otherwise. The categorical kernel used is defined as:
\begin{equation}
    \kappa_h(\bm{h}, \bm{h}') = \frac{\sigma^2}{c} \sum^c_{i=1} 1_\mathbb{C}
    \label{eq:indicator_func}
\end{equation}
where $\mathbb{C}$ is the condition $h=h'$, $c$ is the number of categories for the categorical input and $\sigma^2$ is the kernel variance. We can derive this kernel as a special case of squared exponential kernel~\cite{orr1996introduction}. Recall the standard squared exponential kernel with unit variance being evaluated between two scalar locations $x$ and $x'$:
\begin{equation}
    \kappa(x, x') = \sigma^2 \exp \Big( -\frac{(x - x')^2}{2 \ell^2} \Big)
    \label{eq:unit_sqexp}
\end{equation}

The lengthscale $\ell$ in Eq.~\ref{eq:unit_sqexp} defines the similarity between two input such that as $\ell$ becomes smaller, the distance between locations considered similar (\ie high covariance) shrinks. In the limiting case of $\ell \rightarrow 0$, if two inputs are not exactly the same as each other, then no information would be provided for inferring the GP posterior's value at each other's locations. In other words, the kernel turns into the indicator function shown in Eq.~\ref{eq:indicator_func}:
\begin{equation}
    \kappa(x, x') = 
    \begin{cases}
        1,              & \text{if $x=x'$} \\
        0,              & \text{otherwise}
    \end{cases}
    \label{eq:kernel_cases}
\end{equation}

Thus, for each categorical variable in $\bm{h}$, we can add a squared exponential kernel with $\ell \rightarrow 0$.

\paragraph{Combining kernels}
For combining kernels, we can consider several techniques which result in valid kernels~\cite{duvenaud2013structure}. The most straight-forward approach of summing the categorical and continuous kernels together has been used previously in high dimensional BO and bandits using additive models~\cite{kandasamy2015high}. However, adding the continuous kernel to the categorical kernel (\ie $\kappa_h(\bm{h}, \bm{h}') + \kappa_x(\bm{x}, \bm{x}')$) provided limited expressiveness in the resulting kernel. In practice, this equates to learning a single common trend over $\bm{x}$ with an offset depending on $\bm{h}$. 

Alternatively, we can consider using the product of kernels (\ie $\kappa_h(\bm{h}, \bm{h}') \times \kappa_x(\bm{x}, \bm{x}')$) to improve the expressiveness of the resulting kernel. Multiplicatively combining kernels in this manner allows the encoding of couplings between continuous and categorical domains, which enables a richer set of relationships to be captured. However, during the earlier iterations of BO where there are unlikely to be overlapping categories in the data, the resulting product kernel will be zero which would prevent the model from learning.

\cite{ru2020bayesian} proposed a mixed-kernel as part of their CoCaBO framework which is a mixture of sum and product kernels:
\begin{align}
    \notag \kappa_z(\bm{z}, \bm{z}') &= (1-\lambda)(\kappa_h(\bm{h}, \bm{h}')+\kappa_x(\bm{x}, \bm{x}'))
    \\ &+ \lambda \kappa_h(\bm{h}, \bm{h}')\kappa_x(\bm{x}, \bm{x}')
\end{align}
where a trade-off between the two kernels is controlled by a hyperparameter $\lambda \in [0,1]$ and can be jointly optimised with the GP hyperparameters. For our VPBO method, we adopt the same mixed-kernel as in~\cite{ru2020bayesian}.

\section{Consistency in Performance}
\label{app:consistency_in_performance}
\begin{figure*}[t]
\centering
\subfloat[Func2C]{
\includegraphics[width=0.24\textwidth]{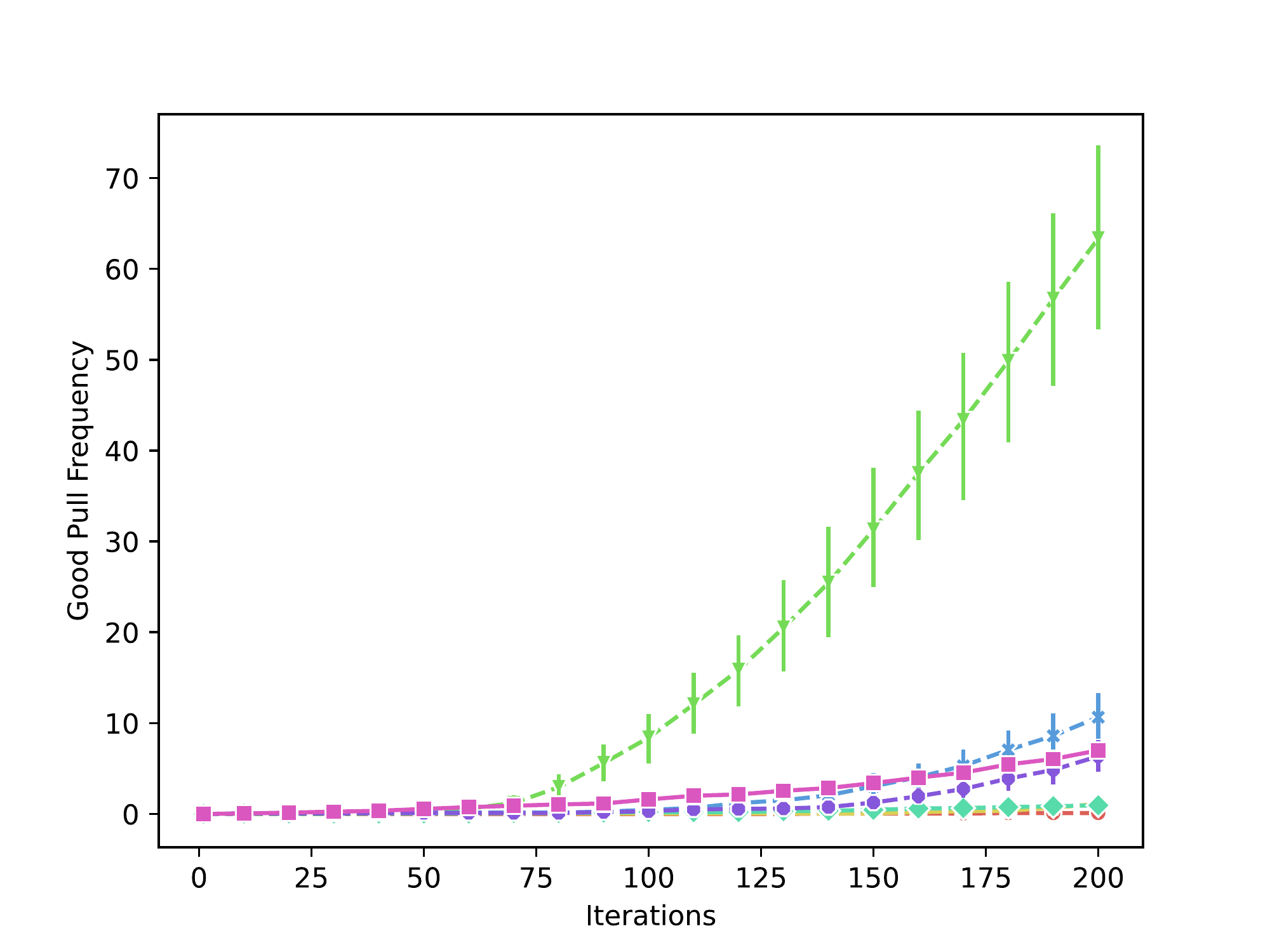}
\label{fig:func2c_valid_arm}
}
\subfloat[Func3C]{
\includegraphics[width=0.24\textwidth]{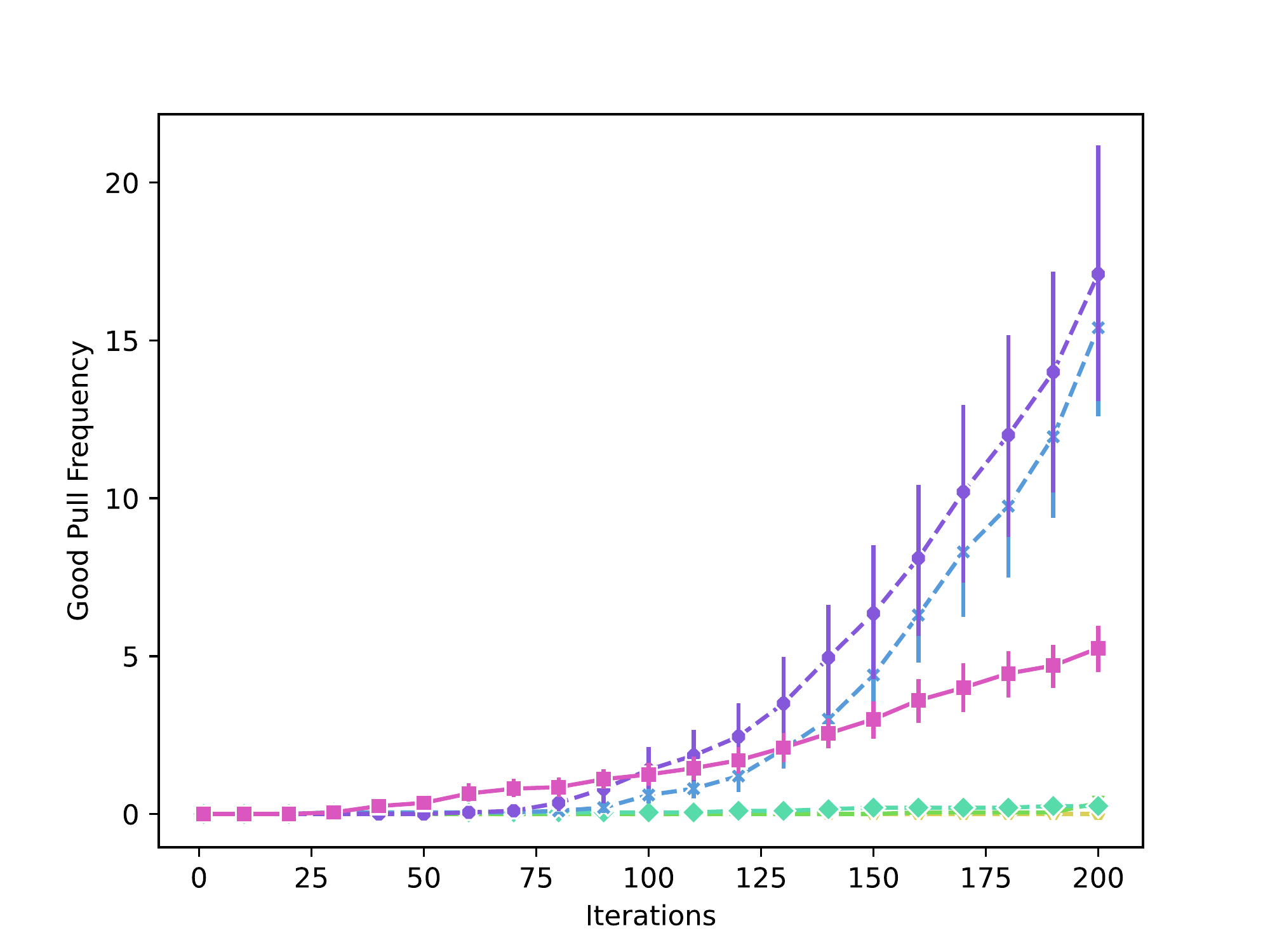}
\label{fig:func3c_valid_arm}
}
\subfloat[Reizman-Suzuki]{
\includegraphics[width=0.24\textwidth]{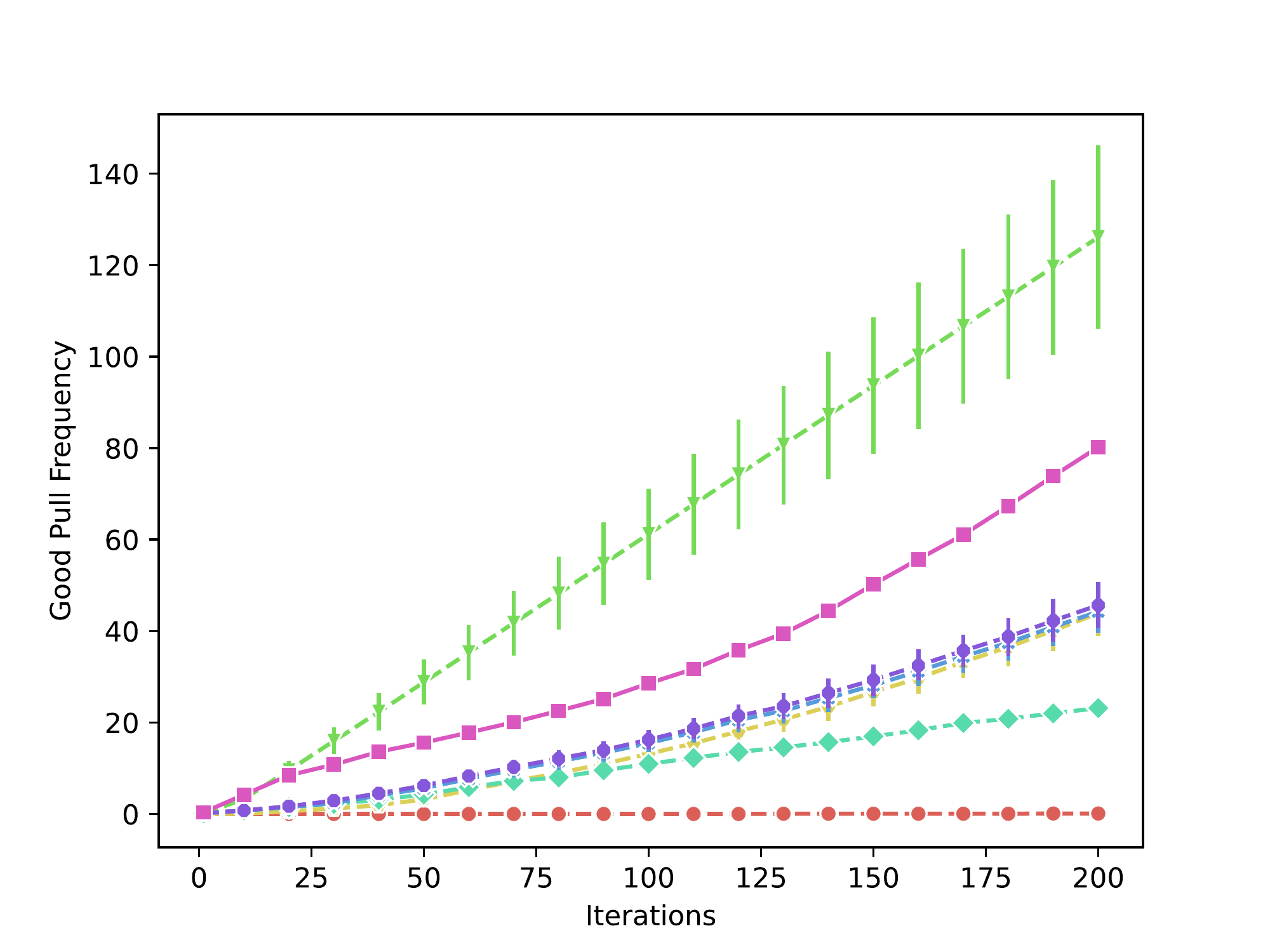}
\label{fig:suzuki_valid_arm}
}
\subfloat[Baumgartner]{
\includegraphics[width=0.24\textwidth]{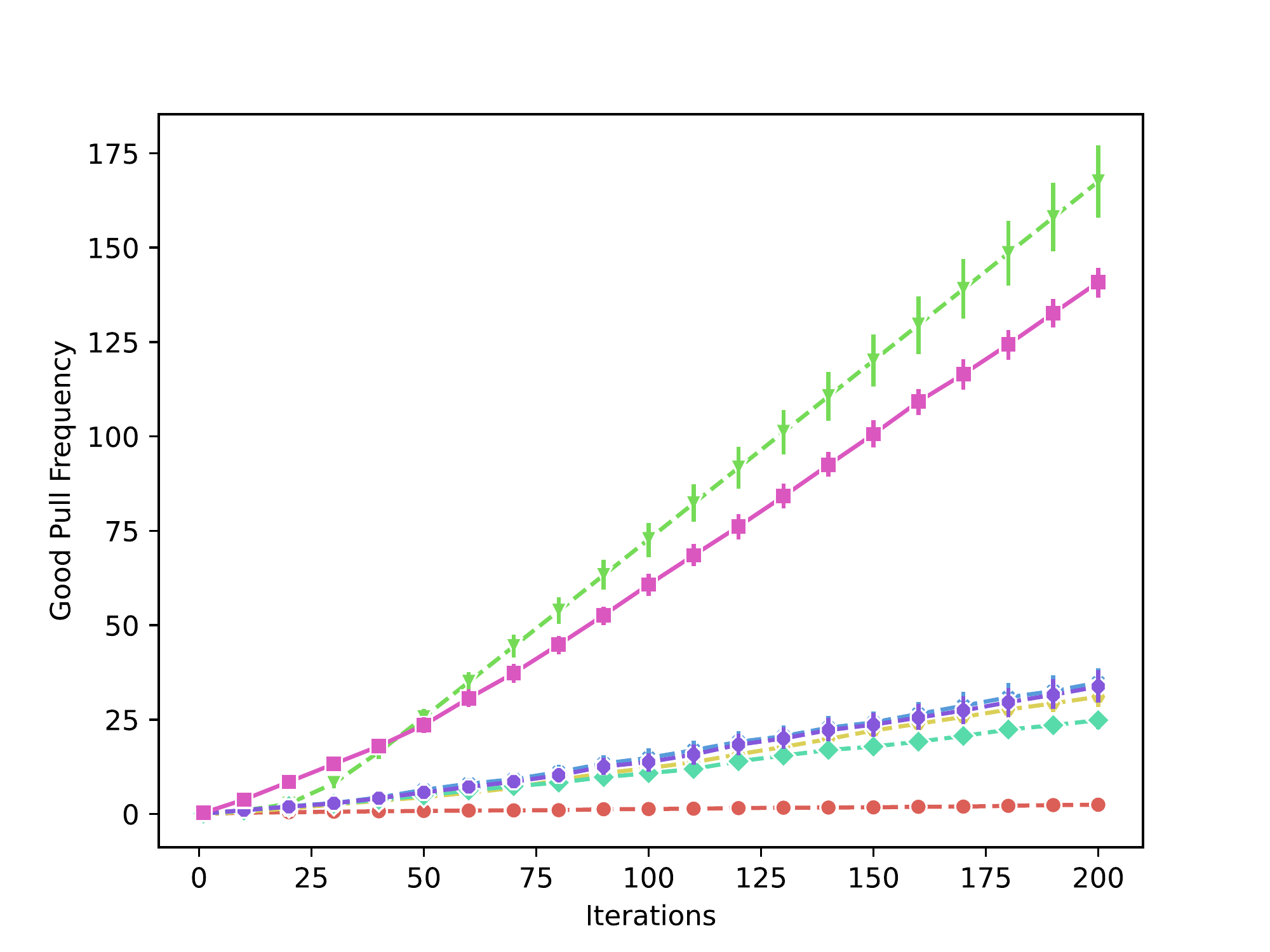}
\label{fig:baumgartner_valid_arm}
} \\
\vspace{-1.0em}
\subfloat[SVM-Boston]{
\includegraphics[width=0.32\textwidth]{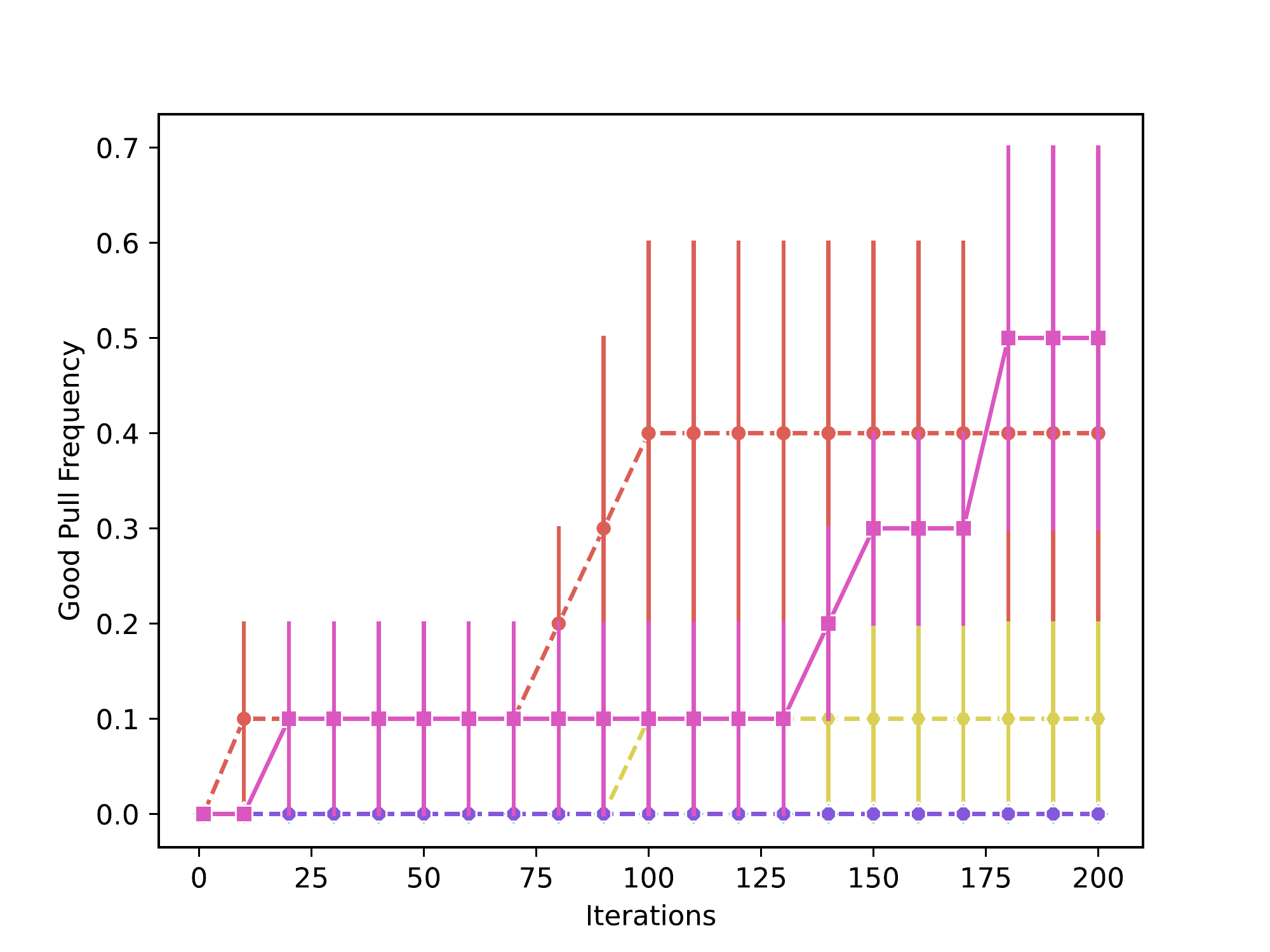}
\label{fig:svmboston_valid_arm}
}
\subfloat[XG-MNIST]{
\includegraphics[width=0.32\textwidth]{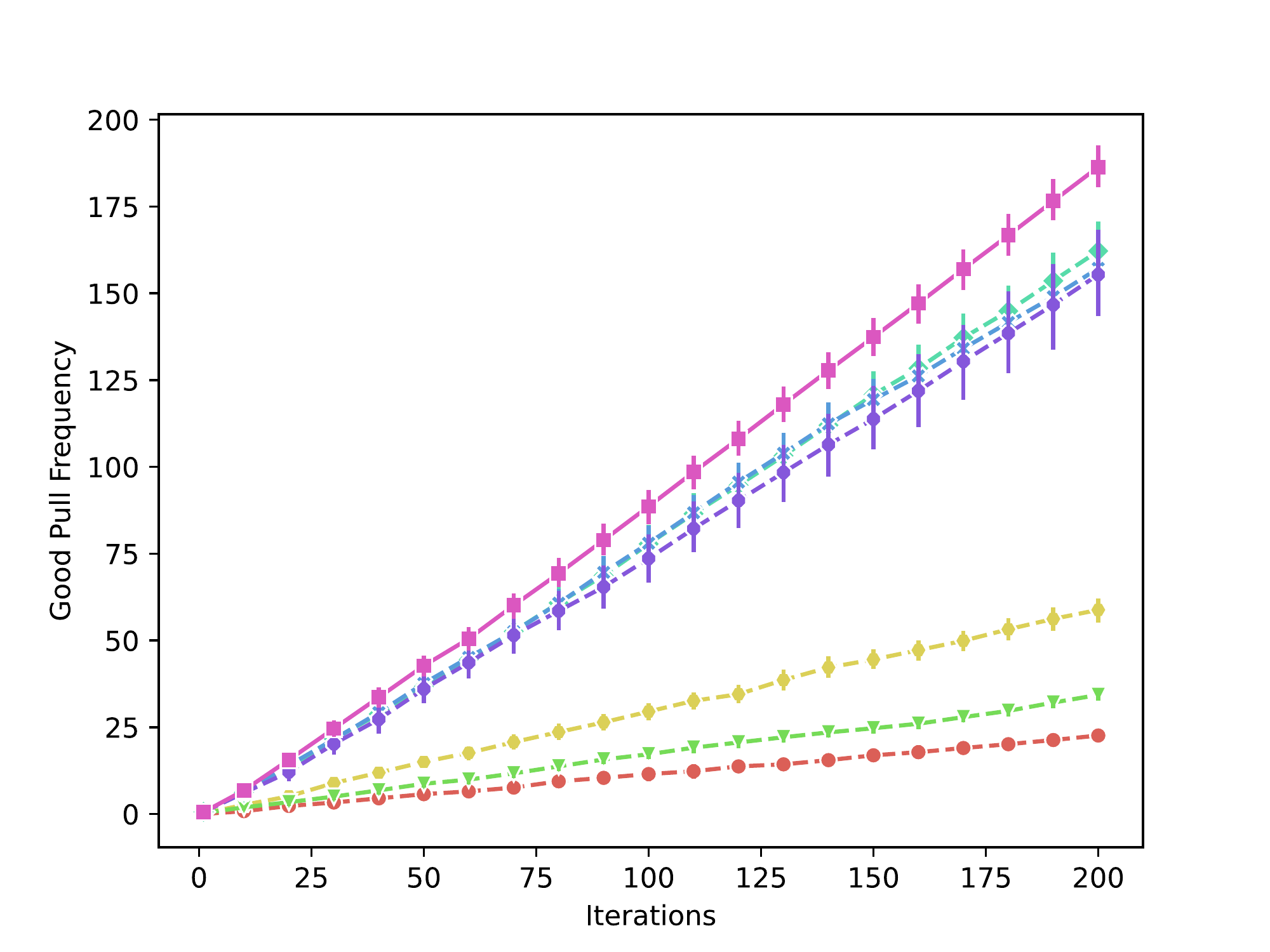}
\label{fig:xgmnist_valid_arm}
}
\subfloat[NASBench-101]{
\includegraphics[width=0.32\textwidth]{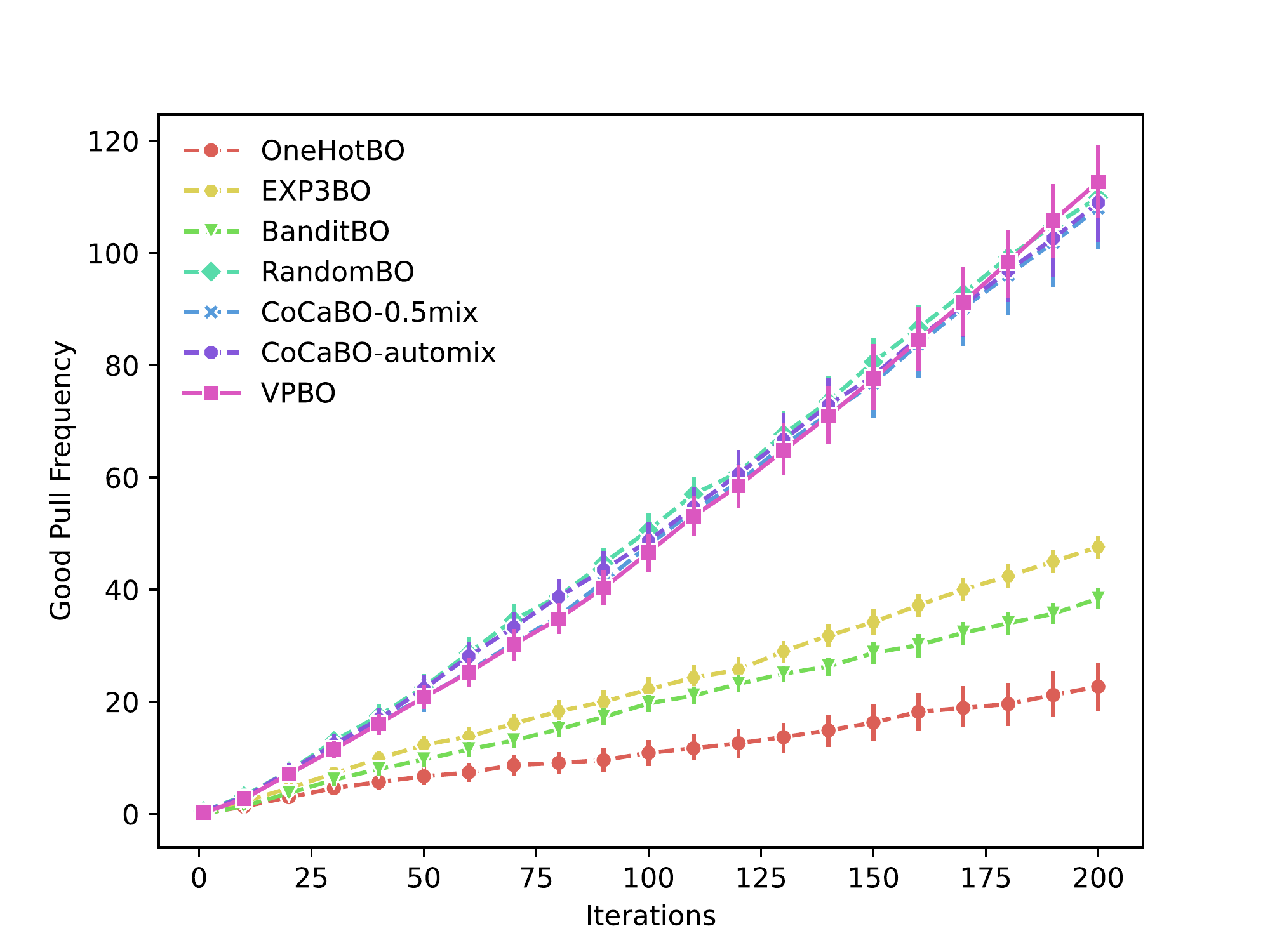}
\label{fig:nasbench101_valid_arm}
}
\caption{Average frequency of making a good choice of categorical variables for several methods on synthetic and real-world tasks.}
\label{fig:valid_arm_plots}
\end{figure*}

Additionally, we inspect the rate at which our VPBO method makes a ``good" choice of categorical variables. Here, we define a good choice as selecting a categorical combination which results in a returned reward which falls within a range of $95\%$ of the ``best" reward as indicated by the Oracle agent. Fig.~\ref{fig:valid_arm_plots} shows the average frequency along with standard error of making a good choice categorical variables over optimisation iterations across $20$ optimisation trials for synthetic datasets and $10$ optimisation trials for real datasets.

Here, we are interested in seeing if each method is able to consistently make good choices for selecting categorical variables which contain high potential for obtaining large rewards. We can see that our VPBO is able to consistently select categorical choices with high potential and obtain higher rewards on average compared to the other baselines. Bandit-BO is able to outperform VPBO across some of the datasets, although we note that Bandit-BO is given a much larger set of initial observations than VPBO. Similarly, CoCaBO is able to outperform our VPBO method on the synthetic datasets, but CoCaBO does not perform very consistently across the multi-objective and real-world datasets.


\section{Wall-clock Time Overhead}
\label{app:wallclock_overhead}
We measure the mean wall-clock time overhead of VPBO compared to several baseline methods, averaged across 200 optimisation iterations. Table~\ref{tab:wallclock_times} shows the average wall-clock time overhead for an optimisation iteration across all $7$ datasets for several methods including VPBO. For each method, the wall-clock time overhead was bench-marked using the same machine. We can see that our VPBO offers comparable wall-clock time overhead with competing baselines and in particular, improved wall-clock time overhead in comparison to the CoCaBO baseline which it is most similar to from a model perspective. Differences in such overhead (a few seconds) is negligible when compared to function evaluation time where the scale of time taken is often in hours or even days.

\begin{table*}[b]
\small
\begin{center}
\begin{tabular}{l c c c c c c}
\specialrule{.2em}{.1em}{.1em}
& \multicolumn{6}{c}{\bfseries Method} \\
\cmidrule(l){2-7}
\thead{Dataset} & \thead{One Hot BO} & \thead{RandomBO} & \thead{EXP3BO} & \thead{BanditBO} & \thead{CoCaBO} & \thead{VPBO} \\
\hline
Func2C & 0.7492 & 0.7525 & 0.3020 & 0.4206 & 0.9600 & 0.5579 \\ 
Func3C & 0.8392 & 0.7629 & 0.2553 & 0.3388 & 0.9118 & 1.0018 \\
Reizman-Suzuki & 0.6861 & 1.0428 & 0.4380 & 0.5951 & 1.1590 & 0.5144 \\
Baumgartner & 1.1952 & 0.8412 & 0.3575 & 0.5056 & 1.0762 & 0.5627 \\
SVM-Boston & 0.6964 & 0.8584 & 0.3550 & 0.3068 & 1.0042 & 0.5504 \\
XG-MNIST & 1.0405 & 3.0324 & 2.4074 & 2.5534 & 3.5881 & 5.0091 \\
NASBench-101 & 0.8771 & 2.2369 & 0.2799 & 10.4477 & 3.2122 & 3.0671 \\
\specialrule{.2em}{.1em}{.1em}
\end{tabular}
\caption{The mean wall-clock time overheads for different methods for each BO iteration across $200$ optimisation rounds in seconds.}
\label{tab:wallclock_times}
\end{center}
\end{table*}

\end{document}